\documentclass[sn-mathphys]{sn-jnl}% Math and Physical Sciences Reference Style
\jyear{2021}%

\theoremstyle{thmstyleone}%
\theoremstyle{thmstyletwo}%

\theoremstyle{thmstylethree}%
\newtheorem{definition}{Definition}%

\usepackage{multirow}
\usepackage{amsmath}
\usepackage[utf8]{inputenc} % allow utf-8 input
\usepackage[T1]{fontenc}    % use 8-bit T1 fonts
\usepackage{hyperref}       % hyperlinks
\usepackage{url}            % simple URL typesetting
\usepackage{booktabs}       % professional-quality tables
\usepackage{amsfonts}       % blackboard math symbols
\usepackage{nicefrac}       % compact symbols for 1/2, etc.
\usepackage{bbding}
\usepackage{float}
\usepackage{microtype}      % microtypography
\usepackage{lipsum}
\usepackage{longtable}
\usepackage{fancyhdr}       % header
\usepackage{graphicx}       % graphics
\usepackage{adjustbox}
\usepackage[utf8]{inputenc}
\usepackage{multicol}  
\usepackage{multirow} 
\usepackage{booktabs}  
\usepackage{threeparttable}  
\usepackage{array}
\graphicspath{{media/}}     % organize your images and other figures under media/ folder
\usepackage{color}
\usepackage{amssymb}
\usepackage{bbm}
\usepackage{xcolor}
\newtheorem{assumption}{Assumption}
\newcommand{\independent}{\protect\mathpalette{\protect\independenT}{\perp}}
\def\independenT#1#2{\mathrel{\rlap{$#1#2$}\mkern2mu{#1#2}}}
\begin{document}

\title[Article Title]{A Survey of Deep Causal Models and Their Industrial Applications}


\author*[1]{\sur{Zongyu Li}}\email{zongyuli@bjtu.edu.cn}

\author*[1,2]{\sur{Xiaobo Guo}}\email{xb\_guo@bjtu.edu.cn}

\author[3]{\sur{Siwei Qiang}}\email{boyue.qsw@mybank.cn}

\affil[1]{\orgdiv{School of Computer and Information Technology}, \orgname{Beijing Jiaotong University}, \city{Beijing}, \postcode{100044}, \country{China}}

\affil[2]{\orgdiv{Data Management Department}, \orgname{China Minsheng Bank}, \city{Beijing}, \postcode{100081}, \country{China}}

\affil[3]{\orgdiv{MYbank}, \orgname{Ant Group}, \city{Beijing}, \postcode{100081}, \country{China}}


\abstract{The notion of causality assumes a paramount position within the realm of human cognition. Over the past few decades, there has been significant advancement in the domain of causal effect estimation across various disciplines, including but not limited to computer science, medicine, economics, and industrial applications. Given the continous advancements in deep learning methodologies, there has been a notable surge in its utilization for the estimation of causal effects using counterfactual data. Typically, deep causal models map the characteristics of covariates to a representation space and then design various objective functions to estimate counterfactual data unbiasedly. {Different from the existing surveys on causal models in machine learning, this review mainly focuses on the overview of the deep causal models based on neural networks, and its core contributions are as follows:} 1) we cast insight on a comprehensive overview of deep causal models from both timeline of  development and method classification perspectives; 2) we outline some typical applications of causal effect estimation to industry; 3) we also endeavor to present a detailed categorization and analysis on relevant datasets, source codes and experiments.}




\maketitle
\resizebox{\textwidth}{!}{
\begin{tabular}{c}
\hline
The authors have no relevant financial or non-financial interests to disclose.\\
\hline 
\end{tabular}
}

\section{Introduction}
In recent years, causal inference has emerged as a popular research topic. In response to various application scenarios, numerous studies adopt diverse approaches to address variation problems. In “The Book of Why” , \cite{TBOW} renowned Turing Award recipient Judea Pearl introduces three levels of causal relationships: association, intervention, and counterfactuals. Pearl's framework advances our understanding of causal relationships by providing a systematic approach to unraveling cause and effect in complex systems.

Currently, the mainstream approaches to causal inference primarily rely on three frameworks: \textbf{Causal Discovery}  \cite{Discovery}, \textbf{Structural Causal Model(SCM) }  \cite{SCM}, and \textbf{Rubin Causal Model(ROM)}  \cite{Rubin}. Table \ref{tab:example} presents a classic example in causal inference known as Simpson's paradox. We have identified a cohort of patients with kidney stones, and from the table, it might be observed that, regardless of whether they belong to the small stone group or the large stone group, treatment A appears to be more effective than treatment B. However, paradoxically, when examining the overall population, treatment B outperforms treatment A. In addressing this issue, causal discovery methods typically aim to identify the causal structure between treatment and outcomes and attempt to identify confounding factors. SCM employ mathematical equations to infer causal relationships between variables and disentangle confounding factors. In contrast to the aforementioned approaches, the ROM predefines the causal structure and utilizes interventions or counterfactual methods to balance confounding interference, making it more convenient for practical applications  \cite{Survey}.

\begin{figure}[H]
    \centering
    \includegraphics[width=0.8\columnwidth]{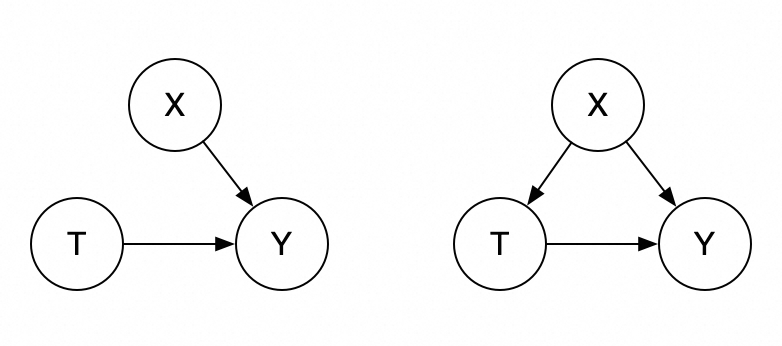}
    \caption{\textbf{Rubin causal model in randomized controlled trial or Observational data}}
    \label{fig:POM}
\end{figure}

In Figure \ref{fig:POM}, the core of the causal framework revolves around the elimination of the direct path between a treatment variable $T$ and an exposure variable $X$, aiming to mitigate the influence of confounding factors, which called \textbf{Causal Effect Estimation}. Currently, two prevailing approaches are widely employed in academic discourse, \textbf{Randomized Controlled Trials(RCT)} and \textbf{Observational Data}. In statistics, we know that RCT are considered the gold standard  \cite{RCT}, but real-world RCT data are sparse and have several serious deficiencies. The studies involving RCT require a large number of samples with little variation in characteristics, which is difficult to interpret and will further involve some ethical issues inevitably. Unlike RCT, observational data, which is collected through observation in natural environments, is easier to obtain and is more widely used in practical applications \cite{causal_observational}.

\begin{table*}[t]
\scriptsize
\renewcommand\arraystretch{1.8}
\renewcommand\tabcolsep{10.0pt} 
\caption{\textbf{A scenario of kidney stone treatment under Simpson's paradox}}
\label{tab:example}
\begin{center}
\vspace{0.2cm}
\resizebox{\textwidth}{!}{
\begin{threeparttable} 
\begin{tabular}{c@{\hspace{1cm}}c@{\hspace{1cm}}c}
\toprule
 \normalsize \textbf{Stone size}& \normalsize \textbf{Treatment A}& \normalsize \textbf{Treatment B}\\
 \hline
 \textbf{Small stones}&  Group 1 \textbf{93\%(81/87)} &  Group 2 87\%(234/270) \\
  \textbf{Large stones}&  Group 3 \textbf{73\%(192/263)} &  Group 4 69\%(55/80) \\
   \textbf{Both}&  78\%(273/350) &   \textbf{83\%(289/350)} \\
\bottomrule
\end{tabular}
\end{threeparttable}
}
\end{center}
\end{table*}

Deep learning largely contributes to the development of artificial intelligence when applied to big data  \cite{Deep_1,Deep_2,Deep_3,Deep_4}. In comparison with traditional machine learning, deep learning models are more computationally efficient, more accurate, and hold significant performance in various fields. However, many deep learning models are black boxes with poor interpretability since they are more interested in correlations than causality as inputs and outputs  \cite{interpretability_1,interpretability_2,explainable}. In recent years, deep learning models have been widely used for mining data for causality rather than correlation  \cite{BNN,CFRNet}. Thus, deep causal models have become a core application for the issue of causal effect estimation in observation data with partial samples  \cite{Deep-Treat,SITE,GANITE,CEVAE}. At present, many works in the field of causal effect estimation utilize deep causal models to select reasonable treatment options  \cite{ACE,DKLITE,DRNet,VCNet}. When encountering observational data, such as whether a patient's blood sugar will decrease after taking medication  \cite{BNN}, deep learning networks might achieve better distribution balance for counterfactual representations of control and treatment groups. They outperform traditional algorithms like BART  \cite{BART} in terms of metrics such as ITE , ATE, and PEHE which can be refer to Section \ref{sec:Examples}.

In recent years, various perspectives have been discussed regarding causal effect estimation  \cite{Survey_0,Survey_1,Survey_2,Survey_3,Survey_4,Survey_5,Survey_6,Survey_7,Survey,Survey_8,Survey_2022}. {In Table \ref{tab:survey}, we list some of existing representative surveys and their highlight points. An in-depth analysis of the origin and development of causal effect estimation was provided in review  \cite{Survey_0}, as well as the implications of causal learning for the development of causal effect estimation. Immediately after that, due to the rapid development of the field of machine learning, a detailed discussion of the relevance of graphical causal effect estimation to machine learning was presented in survey  \cite{Survey_2}. In addition, an overview of traditional and cutting-edge causal effect estimation methods and the comparison between machine learning and causal learning might be found in survey  \cite{Survey_1}. As one of hot focuses in recent years, the studies on the interpretability of machine learning have received great attention. An analytical summary of relevant interpretable artificial intelligence algorithms was elaborated in Kuang. K et al.  \cite{Survey_3} by combining causal effect estimation with machine learning. Moreover, with the flourishing of causal representation learning, review  \cite{Survey_4} takes this new perspective to uncover high-level causal variables from low-level observations, strengthening the link between machine learning and causal effect estimation. In survey  \cite{Survey_8}, the structural causal model of counterfactual intervention was comprehensively explained and summarized, and five classes of problems under causal machine learning are systematically analyzed and compared. Furthermore, in review  \cite{Survey_5}, the authors discussed the way in which the latest progress of machine learning is applied to causal effect estimation, and provides a in-depth interpretation of how causal machine learning might contribute to the advancement in healthcare and precision medicine. As argued in review  \cite{Survey_6}, causal discovery methods might be improved and sorted out based on deep learning, and might also be considered and explored from a variable paradigm perspective. Causal effect estimation in recommender systems is the focus of survey  \cite{Survey_7} , which explains how causal effect estimation might be used to extract causal relationships to enhance recommender systems. The Potential Outcome Framework in statistics has long served as a bridge between causal effect estimation and deep learning. As a starting point, review  \cite{Survey} analyzes and compares variation classes of traditional statistical algorithms and machine learning that satisfy these assumptions. Recently, survey  \cite{Survey_2022} provides an overview of causal learning from Structural Causal Model, Potential Outcome Model, Deep Neural Network and Deep Causal Discovery, explores the internal mechanism of deep learning to improve the significance of causal learning. Different from the above review, we penetrate into the territory of biased sample observation of causal effect estimation, provide a systematic and comprehensive overview of the deep model in representation learning, debias estimation, counterfactual inference and other perspectives.} It is worth mentioning that we provide a comprehensive conclusion of the latest applications of deep causal models in industry. This survey makes three core contributions: 1) we cast insight on a comprehensive overview of deep causal models from both temporal development and method classification perspectives; 2) we outline some typical applications of causal effect estimation to industry; 3) we also try to present a detailed categorization and analysis on relevant datasets, source codes and experiments.

\begin{table*}[t]
\scriptsize
\renewcommand\arraystretch{1.8}
\renewcommand\tabcolsep{10.0pt} 
\caption{\textbf{Highlights of existing surveys on causal learning}}
\label{tab:survey}
\begin{center}
\vspace{0.2cm}
\resizebox{\textwidth}{!}{
\begin{threeparttable} 
\begin{tabular}{c@{\hspace{1cm}}c}
\toprule
 \normalsize \textbf{Surveys}& \normalsize \textbf{Highlights}\\
 \hline
  \cite{Survey_0}& \normalsize The origin and development of causal reasoning\\
  \cite{Survey_2}&\normalsize The connection between graphical causal effect estimation and machine learning\\
  \cite{Survey_3}&\normalsize Machine Learning Interpretability for Counterfactual Causal Reasoning\\
  \cite{Survey_4}&\normalsize Exploring causal variables in data through causal representation learning\\
  \cite{Survey_5}&\normalsize Causal Machine Learning in clinical decision support systems\\
  \cite{Survey_1}&\normalsize The relationship between causal learning and machine learning in big data\\
  \cite{Survey}&\normalsize Causal effect estimation of observational data in Rubin causal Model\\
  \cite{Survey_6}&\normalsize Enhancing causal discovery from the vantage point of deep learning and diverse variable frameworks\\
  \cite{Survey_8}&\normalsize Five types of causal machine learning problems under structural causal models\\
  \cite{Survey_7}&\normalsize Optimize the recommendation system by extracting causality through causal reasoning\\
  \cite{Survey_2022}&\normalsize The significance of deep learning for solving traditional causal learning problems\\
\bottomrule
\end{tabular}
\end{threeparttable}
}
\end{center}
\end{table*}

The rest of the article is outlined as follows. In Section \ref{sec:Development}, the deep causal model is comprehensively elaborated. Next, we categorize the deep causal modeling methods into five groups in Section \ref{sec:Methods}, including learning balanced representation, covariates confounding learning, GANs-based counterfactual simulation  \cite{GAN}, time series causal estimation, and multi-treatment and continuous dose treatment models. Additionally, in Section \ref{sec:Application}, we summarize the industrial applications of causal reference. After that, the relevant experimental guidelines are listed in Section \ref{sec:Experiment}. Finally, we conclude this article in Section \ref{sec:Conclusions}.

\section{Preliminaries}
\label{sec:Basic}
In this section, we present the background knowledge of causal effect estimation, including task descriptions, mathematical notions, and pertinent assumptions.

Basically, the aim of causal effect estimation is to estimate the change in outcome that will occur if a variation treatment are implemented  \cite{Survey}. Imagine that there are several treatment plans A, B, C, and so on, all of which have variation cure rates, and the change in the cure rate is the result of the treatment scheme. Realistically, we unable to apply variation treatment regimens to the same group at the same time. As opposed to RCT, the main problem to be solved in observational research is the lack of counterfactual data. In other words, the challenge we face is how to identify the most effective treatment plan based on past experimental diagnosis and medical history of the patient.

\subsection{Definitions}
Here, to make the survey self-consistent, we first give some basic  definitions under the Potential Outcome Framework  \cite{Rubin}. In particular, causality is defined as the result of a treatment scheme applied to a sample, which might either be a specific behavior, a specific method, or some specific treatment scheme. The followings are the notions related to causal effect estimation, which are benchmarked against the relevant basic definitions in the survey \cite{Survey}.

\begin{definition}
{\textit{Observational data:} Data collected under natural conditions from which confounding factors can be eliminated.}
\end{definition}

{For such data, many popular methods have been developed that remove also confounding from unobservables (using additional assumptions), like instrumental variables, differences-in-differences, and so on.}

\begin{definition}
\textit{Causal Effect Estimation:} Estimate the change in outcome of a given sample receiving an intervention.
\end{definition}
Causal effect estimation and lift modelling have the same goal, the difference is: the lift model is on random experimental data, treatment effect estimates familiarly need to adhere to the necessary assumptions, applied to experimental and observational data. {It is important to note that a causal relationship must be established between a specific intervention and an outcome, rather than merely identifying a spurious association. This implies that the direct effect of the intervention can be isolated from potential confounding variables. Methods such as debiased/double machine learning can also be applied to address the issue of removing spurious correlations in estimating causal effects within deep networks \cite{DML}.} 

\begin{definition}
\textit{Treatment:} A treatment refers to a scheme or action applied to a sample.
\end{definition}
As a medical term, a drug scheme is a treatment. For binary treatments, $T = 1$ is the \textit{treated group}, and $T = 0$ is the \textit{control group}. Multiple treatment might be indicated by the $ T\in \left\{0,1,2,\dots, T_N \right\}$, where $N + 1$ denotes the total number of treatments.

\begin{definition}
\textit{Observed outcome:} An observational outcome, also known as a factual outcome, is a measure of how the sample’s outcomes applied to the treatment.
\end{definition}
In the scenario of a specific treatment, observed outcomes might be displayed in $Y^F$, where $Y^F = Y(T = T_i)$. A donation of in the amount of $Y_{T_i}$ is made as \textit{potential outcome}. 

\begin{definition}
\textit{Counterfactual outcome:} 
A counterfactual outcome is the outcome that differ from a factual outcome.
\end{definition}
With binary treatments, counterfactual outcome is denoted as $Y^{CF}$, and $Y^{CF} = Y(T=1-T_i)$. Assuming multiple treatment, let $Y^{CF}(T=T_i^{'})$ donate the counterfactual result of treatment $T_i^{'}$.
{Which means that there is only one treatment in scenario of binary treatments and in scenario of multiple treatments we have $N-1$ counterfactual outcomes.}

\begin{definition}
\textit{Continuous Treatment:} Continuous Treatment is a consistent and uninterrupted administration of therapeutic interventions or protocols over a specified duration.
\end{definition}
A set of continuous treatment regimens might be donated as $D_T$, the treatment-specific factual treatments might be donated as might be donated as $D^F$, and $D^F = D(T = T_i)$. Simultaneously, counterfactual treatment might be donated as $D^{CF}(T=T_i^{'})$.

\begin{definition}
\textit{Covariates:} Covariates are variables that are unaffected by treatment choice.
\end{definition}
Generally, covariates in the healthcare setting refer to the patient's demographic, medical history, experimental data, etc., familiarly denoted observable confounders in observational data by $X$. However, some confounding factors are unobservable and might be defined as $U$.

\begin{figure}[H]
    \centering
    \includegraphics[width=0.3\columnwidth]{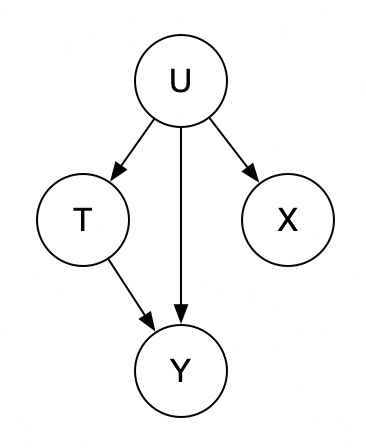}
    \caption{\textbf{Causal diagram of proxy variables under unobserved confounding}}
    \label{fig:CEVAE}
\end{figure}

\begin{definition}
{\textit{Unobservable variables:} Confounding factors that are unobservable in real life.}
\end{definition}
{In the field of socioeconomics, methods that control with specific assumptions for selection on unobservable variables are very popular. For example in Figure \ref{fig:CEVAE}, proxy variables might be used to realize the proxy of unconfounded factors  \cite{CEVAE}. Moreover, for unobserved confounding $U$, one can also commonly employ instrumental variable methods, regression discontinuity design, differences-in-differences approach, as well as synthetic control methods, among others.}

\begin{definition}
{\textit{Instrument Variable:} A contextual element that shapes the treatment decision without exerting a direct influence on the outcome.}
\end{definition}

\begin{figure}[H]
    \centering
    \includegraphics[width=0.5\columnwidth]{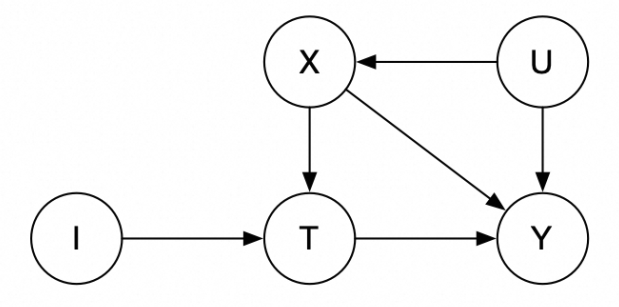}
    \caption{\textbf{Causal diagram of instrumental variables under unobserved confounding}}
    \label{fig:IV}
\end{figure}

According to the illustration in Figure  \ref{fig:IV}, instrumental variables customarily adhere to three essential criteria \cite{IV_Survey}:

\textbf{Relevance:} $I$ is a cause of $T$, i.e., $\mathbb{P}(T \mid I) \neq \mathbb{P}(T)$.

\textbf{Exclusion:} $I$ does not directly affect the outcome $Y$, i.e., $I \independent Y \mid T, \mathbf{X}, \mathbf{U}$

\textbf{Independent:} $I$ is independent of all confounders, including $\mathbf{X}$ and $\mathbf{U}$, i.e., $I \independent \mathbf{X}, \mathbf{U}$

\subsection{Assumptions}
After understanding the basic definition of causal effect estimation, the following three assumptions, which are derived from articles  \cite{Survey,Survey_3}, are commonly required to achieve the estimation of causal treatment effects.

\begin{assumption} {Stable Sample Treatment Value (SSTV):} One sample's response to treatment is independent of the assignment of other samples.
\label{asp: SSTV}
\end{assumption}
Based on this assumption, there is no interaction between samples, and there is only one version of each treatment option. SSTV assumption might be expressed as $P(Y_i \mid T_i,T_i^{'},X_i) = P(Y_i \mid T_i,X_i)$.

\begin{assumption} {Ignorability:}
Given the covariate $X$, the treatment distribution $T$ is independent of the potential outcomes.
\label{asp: Ignorability}
\end{assumption}

On the assumption of ignorability, there should be no unobserved confounding factors. In other words, $T \independent {Y(T = T_i), Y(T = T_i^{'})} \mid X$ needs to be satisfied.

\begin{assumption}
{Overlap:} When given the observed variables, each sample has nonzero probability to receive either treatment status.
\label{asp: Overlap}
\end{assumption}
In order to estimate the counterfactual treatment effect, it must be assumed that each sample might implement any treatment option, otherwise the overlap assumption will not be valid. That is, $0 < P(T = T_i \mid X = x) < 1$ and $0 < P(T = T_i^{'} \mid X = x) < 1$.

\section{Development of Deep Causal Models}
\label{sec:Development}
% In recent years, there is growing dissatisfaction within industry with current machine learning algorithms, 87$\%$ of machine learning projects never make it beyond an experimental phase into production, according to Forbes.
In recent years, there is growing dissatisfaction within industry with current machine learning algorithms, {large mount} of machine learning projects never make it beyond an experimental phase into production, according to Forbes.
However, in the context of big data, with the development of deep learning, the decision-making ability of causal effect estimation has been significantly improved. A recent global survey found that {large mount} of CTOs and senior data scientists agree that causal deep learning might lead to a more robust business environment model, and {large mount} of senior data scientists intend to make a significant investment in deep causal models.
\footnote{\url{https://causalens.com/resources/white-papers/the-causal-ai-revolution-is-happening-now/}}

With a solid understanding of the basic definitions and model measures of causal effect estimation, this section moves to the core of the article. We provide an overview of deep causal models during the last several years and a detailed classification of them.
\subsection{A timeline of development}
\begin{figure}[H]
    \centering
    \includegraphics[width=1\columnwidth]{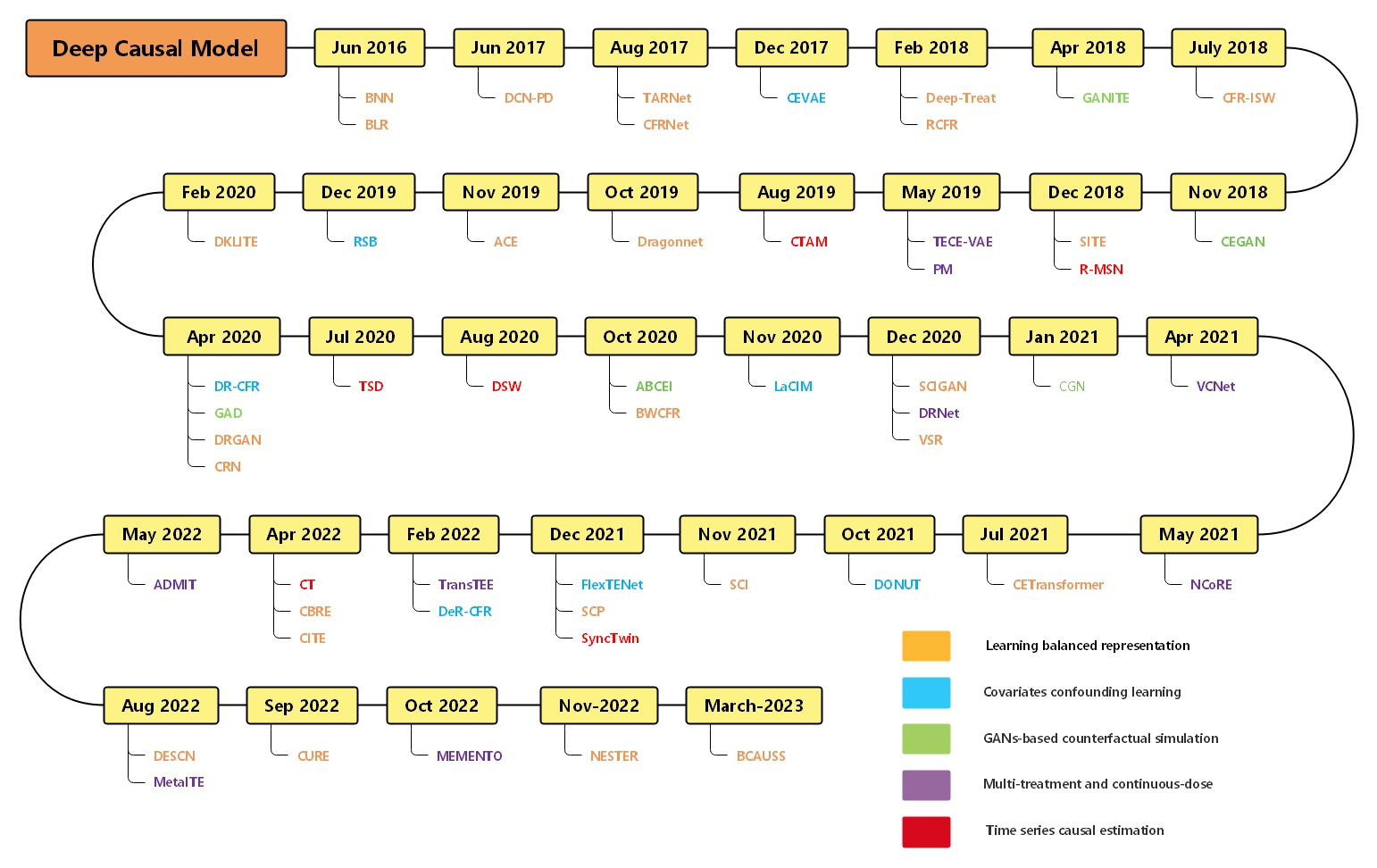}
    \caption{\textbf{Overview of development timeline  of classical deep causal models}}
    \label{fig:process}
\end{figure}

In the past few years, research on deep causal models has advanced considerably, and they have become more accurate and efficient in estimating causal effects. In Figure \ref{fig:process}, we present the development timeline about 50 classical deep causal models from June 2016 to March 2023. It also displays categories of models and the relationship between essays.

Deep causal models have emerged since 2016. For the first time, Johansson et al. publish \textbf{Learning Representations for Counterfactual Inference}  \cite{BNN}, and propose the algorithm framework BNN and BLR  \cite{BNN}, in which  deep learning is combined with the causal effect estimation, and the causal effect estimation problem is transformed into a domain adaptation problem. Since then, a number of models, including DCN-PD  \cite{DCN-PD}, TARNet and CFRNet  \cite{CFRNet} have been proposed. It is important to note that the CEVAE  \cite{CEVAE} model proposed by Louizos et al. in December 2017, which is based on  classical structural Variational Autoencoders, focuses on the impact of confounding factors on the estimation of causal effects.

In 2018, and going forward into 2019, there emerged an increasing interest in causal representation learning with representative works including Deep-Treat  \cite{Deep-Treat} and RCFR  \cite{RCFR} models. After the launch of the GANITE  \cite{GANITE} model, the use of generative adversarial model architecture for counterfactual estimation becomes mainstream in the field of causal effect estimation. In accordance with the previous works, some novel optimization ideas were proposed in CFR-ISW  \cite{CFR-ISW}, CEGAN  \cite{CEGAN}, SITE  \cite{SITE}. 

By applying recurrent neural networks, the R-MSN  \cite{R-MSN} model aims to solve the problem of continuous treatment of multi-treatment time series, which opened up a new theory of deep causal models. Furthermore, PM  \cite{PM} and TECE-VAE \cite{TECE-VAE}, proposed in 2019, attempt to address the issue of estimating causal effects associated with multiple discrete treatment. As a follow-up, the CTAM \cite{CTAM} begins to focus on estimating causal effects for textual data; the Dragonnet \cite{Dragonnet} adhibits regularizations and propensity score networks into causal models for the first time; the ACE \cite{ACE} attempts to extract fine-grained similarity information from representation space. After that, RSB \cite{RSB} model adopts deep representation learning network and PCC regularization for covariate decomposition, uses instrumental variables to control selection bias, and uses confounding factors and moderating factors to predict results.

In 2020, deep causal models got a rapid boom. For DKLITE \cite{DKLITE} model, it  effectively combines the deep kernel trick and posterior variance regularization. Then, DR-CFR \cite{DR-CFR} was proposed to decouple selection bias for covariates; GAD \cite{GAD} focuses on the causal effect of continuous treatment; DRGAN \cite{DRGAN} defines an innovative generative adversarial network for fitting sample dose effect curves; and CRN \cite{CRN} estimates time-varying treatment effects by combining counterfactual recurrent neural networks. Following estimating time series causal effects under multi-cause confounding, TSD \cite{TSD} turns to the estimation of time series causal effects. In the aspect of learning the latent representation space, ABCEI \cite{ABCEI} achieves a significant balance between the covariate distribution of treatment and the control groups by using GAN. As two variants of previous works \cite{CFRNet,CEVAE}, both BWCFR \cite{BWCFR} and LaCIM \cite{LaCIM} are further optimized for the model structure. Additionally, in 2020,  SCIGAN \cite{SCIGAN} and DRNet \cite{DRNet} extended the task of continuous treatment to arbitrary quantity, while VSR \cite{VSR} aggregated the deep latent variables in a reweighted manner.

Up to now since 2021, deep causal models have become more innovative, open, and flexible. The VCNet \cite{VCNet} model implemented an estimator of continuous mean dose-response curves. Moreover, CGN \cite{CGN} proposed more robust and interpretable classifiers that explicitly expose the causal structure of tasks. As for NCoRE \cite{NCoRE}, it used cross-treatment interaction models to explain the process of multi-treatment combination at the potential causal level. After that, inspired by the success application of Transformer in many fields, CETransformer \cite{CETransformer} was proposed to characterize the covariates by focusing the attention on the correlation between the covariates. In addition, DONUT \cite{DONUT} and DeR-CFR \cite{DeR-CFR} are optimized for the two previous works \cite{Dragonnet,DR-CFR} respectively. In  \cite{SCI}, the subspace methods was further explored for causal representation learning, which formulate the SCI model. From the perspective of multi-task learning e, a multi-task adaptive learning architecture was proposed by FlexTENet \cite{FlexTENet}, in which CATE estimation architecture adaptively learns what to share between the PO functions. Aside from that, SCP \cite{SCP} attempted to estimate the multifactorial treatment effects using a two-step procedure. To construct synthetic twin matching representation, SyncTwin \cite{SyncTwin} utilized the temporal structure in the potential outcomes. 

In 2022, TransTEE \cite{TransTEE} extended the distribution balance approach for spatial representations to continuous, structured, and dose-dependent treatments, applying the latest theory of deep learning to expand the depth and breadth of causal effect estimation. Furthermore, DESCN \cite{DESCN} obtaind comprehensive information on treatment propensity, response and hidden treatment effects through a crossover network by means of a multi-task learning. MetaITE \cite{MetaITE} employed a meta-learning counterfactual inference algorithm framework that utilizes information from multiple source treatment groups to perform counterfactual inference under conditions of multiple treatment imbalances. In addition, CBRE \cite{CBRE} employed a cyclic structure to maintain the original data attributes and minimize information loss during the transformation of data into a latent representation space. Immediately after, CITE \cite{CITE} leveraged self-supervised information embedded in the data and achieves a balanced and predictive representation by appropriately utilizing causal prior knowledge. Moreover, ADMIT \cite{ADMIT} provided a generalized bound for estimating ADRF, alleviating selection bias, and offering a discrete approximation of IPM distance with theoretical guarantees. Beside that, CT \cite{CT} combined Transformer with the LSTM network to capture complex long-term dependencies between time-varying confounders. Until recently, CURE \cite{CURE} pre-trained large-scale unlabeled patient data to learn representative background patient representations, and then fine-tunes the labeled patient data for treatment effect estimation. In particular, NESTER \cite{NESTER} utilized neural symbolic programming to address treatment effect estimation problems, allowing for the implicit encoding of inductive bias  within a domain-specific language. Recently, BCAUSS \cite{BCAUSS} had achieved the characterization of generalization error and the distance representation between the control group and the treatment group through automated balanced networks and self-supervised learning.

%The next subsection provides a detailed analysis of classical deep causal models and a categorical comparison of network structures and implementation ideas.

\subsection{Model classification}
\begin{figure}[H]
    \centering
    \includegraphics[width=1\columnwidth]{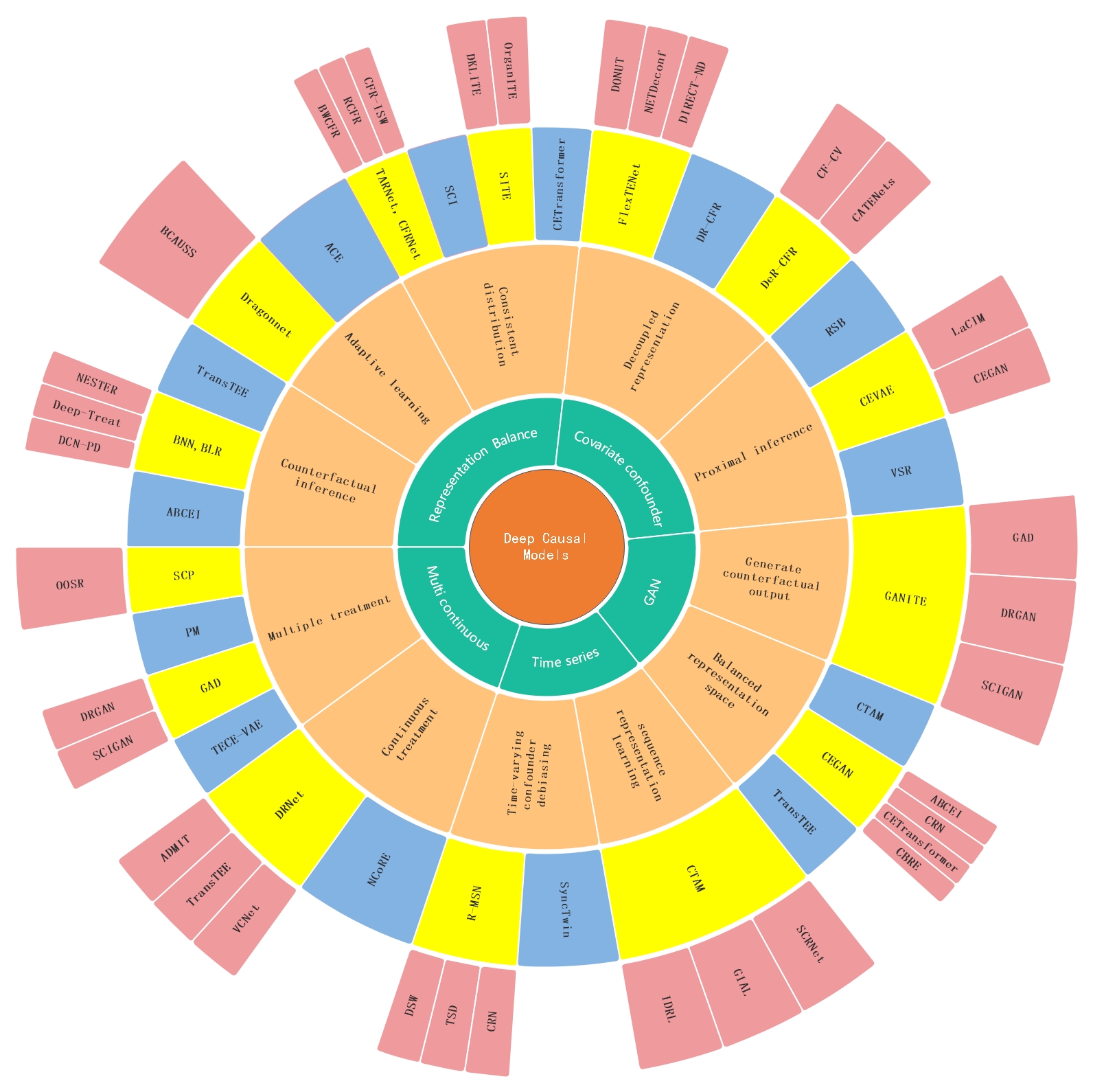}
    \caption{\textbf{{Taxonomy of deep causal models}}}
    \label{fig:Taxonomy}
\end{figure}

Through sorting out the existing models, we might identify that the current deep causal models are mainly studied from the following aspects that include : 1) Learning balanced representations; 2) Covariate confounding learning; 3) Time series causal learning; 4) GANs based Counterfactual simulation; and 5) Multi-treatment and continuous treatment. In Figure \ref{fig:Taxonomy}, we present a detailed classification of the current deep causal model.

%Considering the perspectives from representing the distributional balance of spatial factual outcomes and counterfactual outcomes, unbiased estimation learning with covariate decoupling and de-confounding, simulation generation of counterfactual outcomes and spatial representations by generative adversarial networks, estimation of temporal causal situations by recurrent neural networks, application of multiple treatments as well as continuous dose treatment of attention mechanisms, etc. We divided the existing models into five categories: 1) Learning balanced representations methods; 2) Covariates confounding learning methods; 3) Methods based on Generative Adversarial Networks; 4) Time series causal estimation problem; 5) Methods based on multi-treatment and continuous dose models.
\begin{itemize}
    \item {\textbf{Learning
    balanced representations:} {This type of approach has long been a popular research. The core ideology is to use the encoder to map the covariates $X$ to the representation space $\Phi$, combine the processing $T$, adopt the network $h$ to predict the output outcome $Y$, and minimize the distribution distance $\operatorname{disc}_{\mathcal{H}}$ between the factual and counterfactual outcomes. Additionally, adopting adaptive learning network structures can enhance the model's generalization capabilities, and counterfactual inference prediction networks can be implemented to achieve representation of unbiased estimates. Classical architectures include BNN \cite{BNN}, CFRNet \cite{CFRNet}, Dragonnet \cite{Dragonnet}, SITE \cite{SITE}, ACE \cite{ACE}, DKLITE \cite{DKLITE}, SCI \cite{SCI}.}}
    
    \item {\textbf{Covariate confounding learning:} {This type of approach aims to decomposition of covariant relation in theory. Its main application schemes are unbiased estimation of covariates and removal of confounding factors using decoupling, reweighting, codec reconstruction, etc. For unobserved confounders, causal effects can be substituted using surrogate reasoning, regression discontinuity design, differences-in-differences, or synthetic cohorts. The typical structures include CEVAE \cite{CEVAE}, DeR-CFR \cite{DeR-CFR}, LaCIM \cite{LaCIM}, DONUT \cite{DONUT}, FlexTENet \cite{FlexTENet}.}}

    \item {\textbf{Time series causal estimation:} {Temporal causal estimation has been widely concerned. Using RNNs to track contextual covariate information and handle time-varying confounding bias is a long-standing solution adopted by many models. Utilizing sequential representation network structures allows for better exploitation of historical sequence data, enabling subsequent inference of medical conditions. Some typical architectures are R-MSE \cite{R-MSN}, CTAM \cite{CTAM}, CRN \cite{CRN}, TSD \cite{TSD}.}}
    
    \item {\textbf{GANs-based counterfactual simulation:} {With the great success of GANs in data synthesis in recent years, it is also widely adopted to solve causal effect estimation problems. Two schemes are familiarly involved in Using GAN networks for counterfactual simulation, i.e., generating the counterfactual output outcomes or balanced representation space distributions. The discriminator structure of GAN networks significantly improves the reliability of counterfactual data, while achieving consistent distribution between experimental and control groups. Such classical frameworks include GANITE \cite{GANITE}, CEGAN \cite{CEGAN}, ABCEI \cite{ABCEI}, CETtrnaformer \cite{CETransformer}.}}
    
    \item {\textbf{Multi-treatment and continuous treatment:} {The issues of Multiple treatment and continuous treatment are one of recent research hotspots in deep causal learning. In general, such issues might be further simplified and structured using schemes such as matching, variational autoencoders, hierarchical discriminators, and multi-headed attention mechanisms. In addition, innovative network structures combine multiple and continuous treatment issues, leading to flexible structured counterfactual predictions. The classical models include PM \cite{PM}, TECE-VAE \cite{TECE-VAE}, DRNet \cite{DRNet}, SCIGAN \cite{SCIGAN}, VCNet \cite{VCNet}, TransTEE \cite{TransTEE}.}}
    
\end{itemize}

It is worth pointing out, as variation application scenarios arise, new research focuses and methods in the aspect of deep causality model will also emerge continuously. In the following section, we will give a detailed introductions to the typical models according to the taxonomy of deep causal models motioned above.

\section{Typical Deep Causal Models}
\label{sec:Methods}

\begin{table*}[ht]
\scriptsize
\renewcommand\arraystretch{1.6}
\renewcommand\tabcolsep{25.0pt} 
\caption{\textbf{Highlights of deep frameworks on classical causal models}} 
\label{tab:model}
\begin{center}
\resizebox{\textwidth}{!}{
\begin{threeparttable} 
\begin{tabular}{c@{\hspace{1cm}}c@{\hspace{1cm}}c@{\hspace{1cm}}c@{\hspace{1cm}}c@{\hspace{1cm}}c@{\hspace{1cm}}c}
\toprule
 \normalsize \textbf{Models} & \normalsize \textbf{GAN} & \normalsize \textbf{AE} & \normalsize \textbf{RNN} & \normalsize \textbf{Transformer} & \normalsize \textbf{Citations} \\
  \hline
%DCN-PD \cite{DCN-PD}&&&&\\
 \normalsize BNN \cite{BNN}&&\Checkmark&&&\normalsize 685\\
 \normalsize CFRNet \cite{CFRNet}&&\Checkmark&&&\normalsize 871\\
 \normalsize CEVAE \cite{CEVAE}&&\Checkmark&&&\normalsize 676\\
% Deep-Treat \cite{Deep-Treat}&&\Checkmark&&&\\
% RCFR \cite{RCFR}&&\Checkmark&&\\
 \normalsize GANITE \cite{GANITE}&\Checkmark&&&&\normalsize 354\\
% CFR-ISW \cite{CFR-ISW}&&\Checkmark&&\\
% CEGAN \cite{CEGAN}&\Checkmark&&&\\
 \normalsize SITE \cite{SITE}&&\Checkmark&&&\normalsize 256\\
 \normalsize R-MSN \cite{R-MSN}&&&\Checkmark&&\normalsize 110\\
 \normalsize PM \cite{PM}&&\Checkmark&&&\normalsize 97\\
 \normalsize TECE-VAE \cite{TECE-VAE}&&\Checkmark&&&\normalsize 14\\
 \normalsize CTAM \cite{CTAM}&\Checkmark&\Checkmark&&&\normalsize 23\\
 \normalsize Dragonnet \cite{Dragonnet}&&\Checkmark&&&\normalsize 275\\
 \normalsize ACE \cite{ACE}&&\Checkmark&&&\normalsize 22\\
 %RSB \cite{RSB}&&\Checkmark&&\\
 \normalsize DKLITE \cite{DKLITE}&&\Checkmark&&&\normalsize 84\\
%  DR-CFR \cite{DR-CFR}&&\Checkmark&&\\
 \normalsize GAD \cite{GAD}&\Checkmark&&&&\normalsize 15\\
 %DRGAN \cite{DRGAN}&\Checkmark&\Checkmark&&\\
 \normalsize CRN \cite{CRN}&\Checkmark&&\Checkmark&&\normalsize 130\\
 \normalsize TSD \cite{TSD}&&&\Checkmark&&\normalsize 73\\
 \normalsize ABCEI \cite{ABCEI}&\Checkmark&\Checkmark&&&\normalsize 32\\
% BWCFR \cite{BWCFR}&&\Checkmark&&\\
 %LaCIM \cite{LaCIM}&&\Checkmark&&\\
 \normalsize SCIGAN \cite{SCIGAN}&\Checkmark&\Checkmark&&&\normalsize 64\\
 \normalsize DRNet \cite{DRNet}&&\Checkmark&&&\normalsize 94\\
 %VSR \cite{VSR}&&\Checkmark&&\\
 \normalsize VCNet \cite{VCNet}&&\Checkmark&&&\normalsize 38\\
 %NCoRE \cite{NCoRE}&&\Checkmark&&\\
 \normalsize CETransformer \cite{CETransformer}&\Checkmark&\Checkmark&&\Checkmark&\normalsize 9\\
 \normalsize DeR-CFR \cite{DeR-CFR}&&\Checkmark&&&\normalsize 38\\
 \normalsize SCI \cite{SCI}&&\Checkmark&&&\normalsize 7\\
 %WUNT \cite{WUNT}&&&&\Checkmark\\
 \normalsize FlexTENet \cite{FlexTENet}&&\Checkmark&&&\normalsize 42\\
 \normalsize SCP \cite{SCP}&&\Checkmark&&&\normalsize 10\\
 \normalsize CGN \cite{CGN}&\Checkmark&\Checkmark&&&\normalsize 103\\
 \normalsize SyncTwin \cite{SyncTwin}&&&\Checkmark&&\normalsize 18\\
 \normalsize TransTEE \cite{TransTEE}&\Checkmark&\Checkmark&&\Checkmark&\normalsize 9\\
 %DESCN \cite{DESCN}&&\Checkmark&&&\\
 %CURE \cite{CURE}&&\Checkmark&&\Checkmark&1\\
 \normalsize DSW \cite{DSW}&&\Checkmark&\Checkmark&&\normalsize 14\\
 %CITE \cite{CITE}&&\Checkmark&&\\
 \normalsize CBRE \cite{CBRE}&\Checkmark&\Checkmark&&&\normalsize 9\\
 \normalsize CT \cite{CT}&\Checkmark&\Checkmark&\Checkmark&\Checkmark&\normalsize 31\\
 \bottomrule
\end{tabular}
\end{threeparttable}
}
\end{center}
\end{table*}

Deep learning has witnessed increasing utilization across diverse domains, including healthcare, education, and economy, for discerning causal connections from counterfactual data. Unlike conventional deep causal models that map covariates to a representation space, the objective function might provide a more impartial estimation of counterfactual data. Unlike the brief overview of the various classical models that are categorized from the variation research perspectives for deep causal models, Table \ref{tab:model} summarizes the classical network architecture applied by those typical deep causal models. In addition, in the  following detailed descriptions on the typical deep learning-based causal models, the issues and challenges that these models face are also discussed.

\subsection{Learning balanced representation}
In reality, test and training data often have related but not identical distributions, which is a challenge for statistical learning theories. To solve this problem in the field of causal effect estimation, deep learning models that learn causality instead of correlation are used. Unlike RCT \cite{RCT}, deep causal models are useful even if we have only train observational datasets, means that we might estimate ATEs or ITEs in the train-set only. Therefore, to predict counterfactual outcomes by learning from factual data, causal effect estimation needs to be transformed into a domain adaptation problem \cite{BNN}.

\begin{figure}[ht]
    \centering
    \includegraphics[width=0.62\columnwidth]{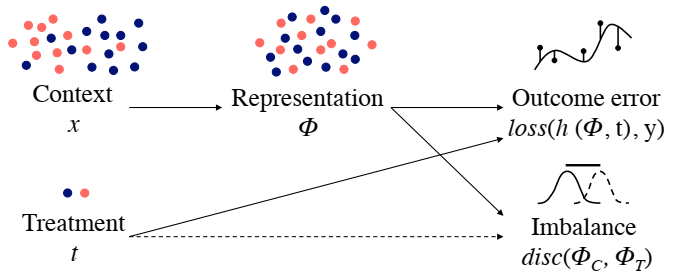}
    \caption{\textbf{Balancing Neural Network (BNN) }: Counterfactual inference networks based on representation distribution balance \cite{BNN}}
    \label{fig:Counterfactual}
\end{figure}

For counterfactual results to be predicted, effective feature representations are necessary, especially the balanced distributions. According to Johansson et al, BNN \cite{BNN} is an algorithmic framework as shown in Figure \ref{fig:Counterfactual} for counterfactual reasoning that transforms the causal effect estimation problem into a representation distribution balance problem.

Upon mapping the covariates to the representation space, the encoder makes use of a two-layer fully connected neural network, balances the distribution distance of the representation space, and then derives the counterfactual results using another two-layer fully connected network. The used regression function is as follows:
\begin{equation}
\resizebox{.9\hsize}{!}{ $ B_{\mathcal{H}, \alpha, \gamma}(\Phi, h)=\frac{1}{n} \sum_{i=1}^{n}\mid h\left(\Phi\left(x_{i}\right), t_{i}\right)-y_{i}^{F} \mid+\alpha \operatorname{disc}_{\mathcal{H}}\left(\hat{P}_{\Phi}^{F}, \hat{P}_{\Phi}^{C F}\right)+\frac{\gamma}{n} \sum_{i=1}^{n} \mid h\left(\Phi\left(x_{i}\right), 1-t_{i}\right)-y_{j(i)}^{F} \mid $ },
\label{equ:BNN}
\end{equation}
where the encoder network is denoted by $\Phi$, a predictor network by $h$, and the metric function is $\operatorname{disc}_{\mathcal{H}}$ that represents the distance between the two distributions. Aside that, \ref{equ:BNN} minimizes the error of the training set facts. 

As an innovative method for measuring the spatial distribution distance between treatment groups and control groups, the literature \cite{CFRNet} proposed a CFRNet network structure based on BNN \cite{BNN} and adopted MMD and WASS for spatial distribution distance representations. When the network is trained, the imbalance penalty is calculated based on the explicit boundary of the distance, and the loss is calculated separately for the treatment group and the control group. As well as adding multiple neural network layers between results predition layer, DCN-PD \cite{DCN-PD} combined multi-task deep neural networks with propensity score dropout.

On the basis of the CFRNet \cite{CFRNet} model, both RCFR \cite{RCFR} and CFR-ISW \cite{CFR-ISW} used the Propensity score to re-weight the representative spatial feature region and the sampling objective function; Atan et al. proposed an unbiased autoencoder network Deep-Treat \cite{Deep-Treat} framework, applying a feedforward neural network to learn the optimal treatment strategy. While it reduces the loss of representation reconstruction as well as the information loss in space, the selection bias is also narrowed. 

\begin{figure}[H]
    \centering
    \includegraphics[width=0.6\columnwidth]{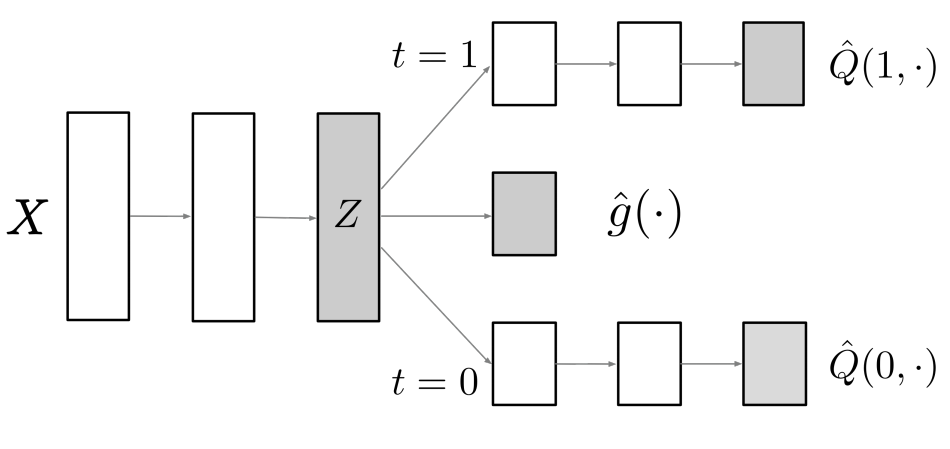}
    \caption{\textbf{Dragonnet}: Propensity score adaptive neural network \cite{Dragonnet}}
    \label{fig:Dragonnet}
\end{figure}

Proposed by Shi et al., first-ever Dragonnet \cite{Dragonnet} adhibited the regularization objective functions into nonparametric estimation theory and the Propensity score prediction networks to CFRNet \cite{CFRNet}, thus making sure that the covariates are adjusted for treatment-related information in them. As might be seen, Figure \ref{fig:Dragonnet} shows the network structure of the propensity score adaptive neural network. 

To maintain local similarity and balance of data representing the treatment and control groups and to improve individual treatment outcomes. Yao et al. proposed the SITE \cite{SITE} method, which combines position-dependent depth metric PDDM with midpoint distance minimization MPDM into the representation space, and predicts the potential outcomes using a binary result network. In this scenario, the following loss function is used:
\begin{equation}
\mathcal{L}=\mathcal{L}_{\mathrm{FL}}+\beta \mathcal{L}_{\mathrm{PDDM}}+\gamma \mathcal{L}_{\mathrm{MPDM}}+\lambda\|W\|_{2},
\end{equation}
{it employs various techniques to minimize errors and prevent overfitting, such as $\mathcal{L}_{\mathrm{FL}}$ for comparing predicted and actual factual results, $\mathcal{L}_{\mathrm{PDDM}}$ and $\mathcal{L}_{\mathrm{PDDM}}$ for evaluating the MPDM and PDDM, and ${L}_2$ regularization for the parameter $W$.}
Here, the position-dependent deep component gauges the local similarity between two units based on their relative and absolute positions in the latent space $Z$. Conversely,
the middle point distance minimization aims to balance the representation distribution between treated and control groups by minimizing the distances to the midpoints between unit points.

According to SITE \cite{SITE}, ACE \cite{ACE} proposed a balanced and adaptive similarity regularization structure to extract spatially fine-grained similarity information; in DKLITE \cite{DKLITE} was proposed by applying the deep kernel regression and a posterior regularization to learn the spatial domain overlap information; and BWCFR \cite{BWCFR} re-weighted the spatial feature distribution for the domain overlap region. In addition, CBRE \cite{CBRE} utilized cyclically balanced representation learning to retain crucial information about the original data attributes and minimize information loss when converting data into a latent representation space during counterfactual inference; CITE \cite{CITE} constructed a framework for comparing individual treatment effects by leveraging self-supervised information hidden in the data, achieving balanced and predictive representations while appropriately utilizing causal prior knowledge.

Unlike the above models, many other works attempt to combine representation distribution balancing with other deep network models. By combining GANs with a mutual information estimator regularization structure, ABCEI \cite{ABCEI} tries to balance the covariate distributions of the treatment and control groups in the representation space; CETransformer \cite{CETransformer} was proposed to apply the attention mechanism to focus on the relationships between covariates and then learn a balanced representation distribution; as TransTEE \cite{TransTEE} extends the balanced representation distribution method to Continuous, Structured, and Dose-Related treatments, it makes causal effect estimation a more open-ended problem; CURE \cite{CURE} designed a new sequence encoding for longitudinal (or structured) patient data and merge structure and time into patient embeddings; DESCN \cite{DESCN} learned the processing and response functions jointly across the sample space to avoid processing bias, and used an intermediate pseudo-processing effect prediction network to mitigate sample imbalance.

\begin{figure}[H]
    \centering
    \includegraphics[width=0.72\columnwidth]{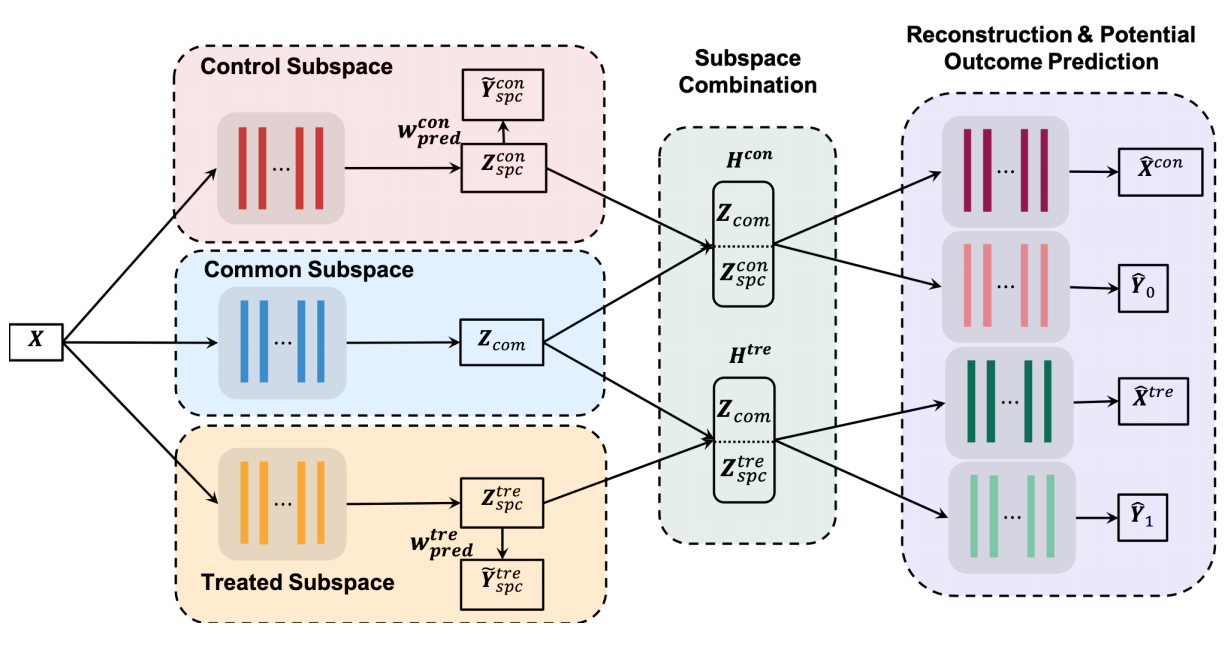}
    \caption{\textbf{Subspace Learning Based Counterfactual Inference network (SCI)}: Estimating individual causal effects while preserving information about specific and common subspaces \cite{SCI}}
    \label{fig:SCI}
\end{figure}

It is worth mentioning that SCI \cite{SCI} inducted the notion of a subspace as shown in Figure \ref{fig:SCI}, integrating the covariates into a common subspace, a treatment subspace, and a control subspace simultaneously, thereby obtaining a common representation and two specific representations. Afterwards, the common representation is connected to the specific representation of the treatment group and of the control group, and two potential results are obtained from the reconstruction and prediction network. Based on SCI, NETDECONF \cite{NETDECONF} used network structure information to infer hidden confounding in the observational data. A personalized treatment effect model that allocates treatments according to scarcity and estimates potential outcomes is proposed by OrganITE \cite{OrganITE}. It is worth mentioning that NESTER \cite{NESTER} employed inductive bias and heuristics to design multi-head neural network structures and regularizers, and applied neural symbolic programming instead of traditional methods for treatment effect estimation. Recently, BCAUSS \cite{BCAUSS} utilized a self-supervised automated balancing covariate representation method to minimize the generalization error and reduce the distance between the control and treatment groups.

Due to the improvement in the feasibility of estimating causal effect, the representation distribution balance is becoming the mainstream way, but it is still limited to estimating individual treatment effects, and thus is hard to expand to a broader range of applications like multi-treatments and continuous treatments.
\subsection{Covariates confounding learning}
The main issue in causal effect estimation is estimating the treatment effect when given a covariate, a treatment, and a predicted outcome. By identifying and correcting for confounders, it is possible to estimate causal effects with greater accuracy from observational data. Nevertheless, in practical scenarios, there are potential confounders of noise and uncertainty, as well as some non-confounders. For this reason, mining potential confounders and decoupling the associated covariates are important methods to learn unbiased representations of counterfactuals from observed data.

A CEVAE \cite{CEVAE} model structure was first proposed by Louizos et al. to capture hidden confounding factors with VAEs in the presence of noise and uncertain confounding factors, and perform treatment and prediction. On the basis of TARNet \cite{CFRNet}'s causal relationship diagram structure, Do-calculus derivations are performed for $y$ and $t$ in the inference network, respectively, and $z$ and $t$ in the model network to fit the interaction between potential confounding variables and treatment effects. Overall, the following prediction function is involved in the causal variational autoencoder: 
\begin{equation}
\mathcal{F}_{\text {CEVAE }}=\mathcal{L}+\sum_{i=1}^{N}\left(\log q\left(t_{i}=t_{i}^{*} \mid \mathbf{x}_{i}^{*}\right)+\log q\left(y_{i}=y_{i}^{*} \mid \mathbf{x}_{i}^{*}, t_{i}^{*}\right)\right),
\end{equation}
the training set comprises observations of an input variable, treatment variable, and outcome variable at specific data points $\mathbf{x}_{i}^{*}$, ${t}_{i}^{*}$, and ${y}_{i}^{*}$. {In this context, $q(\mathbf{z}, t, y \mid \mathbf{x})$ represents the inference network, while $z$ denotes the unobserved confounding factors.}

On the ground of CEVAE \cite{CEVAE}, Sun et al. proposed the LaCIM \cite{LaCIM} latent causal model to avoid false associations and improve the generalization ability of the model; CEGAN \cite{CEGAN} used GAN networks to identify the potential confounders unbiasedly.

In accordance with CEVAE \cite{CEVAE}, VSR \cite{VSR} proposed a reweighting model that removes the association processing and confounding factors. It also used a deep neural network to aggregate the density ratios of latent variables across the variational distribution for calculating the sample weight distribution; as part of the estimation process, DONUT \cite{DONUT} has added an orthogonal constraint to the non-confounding factors in the total loss; in addition, an end-to-end regularization and reparameterization method called FlexTENet \cite{FlexTENet} was proposed to learn a new architecture using multi-tasking framework, by which  the shared function between causal structures is obtained adaptively. In DIRECT-ND \cite{DIRECT-ND}, the entangled representation is solved through hybrid learning, and the multivariate causal effect estimation problem was studied from a new perspective. Additionally, VAE and GAN networks were added to the model to realize the learning of hybrid representation space.

Aiming at balancing selection bias, Zhang et al.  \cite{RSB} proposed the RSB algorithm using autoencoder networks via PCC regularization and instrumental variables, and also adding the confounding variables and moderators for prediction. As extention of CFRNet \cite{CFRNet},  both DR-CFR \cite{DR-CFR} and DeR-CFR \cite{DeR-CFR} were proposed with the main structure as shown in Figure \ref{fig:DeR-CFR} to remove the covariate correlations. 

As depicted in Figure \ref{fig:DeR-CFR}, the observed covariates $X$ might be influenced by three distinct factors: $I$, an instrumental factor that solely affects the treatment $T$; $C$, a confounding factor that impacts both the treatment $T$ and the outcome $Y$; and $A$, an adjustment factor that determines the outcome $Y$. To learn the decomposed representation for counterfactual reasoning, the following steps are taken \cite{DeR-CFR}:

\begin{figure}[H]
    \centering
    \includegraphics[width=0.35\columnwidth]{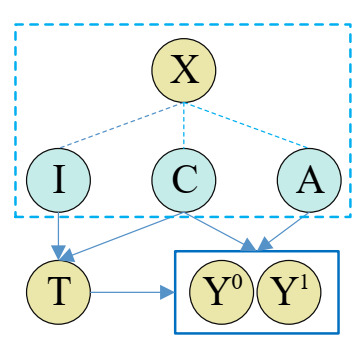}
    \caption{\textbf{Decomposed Representation for CounterFactual Regression (DeR-CFR)}: Causal framework with decomposed latent factors \cite{DeR-CFR}}
    \label{fig:DeR-CFR}
\end{figure}

\begin{itemize}
  \item Representation networks for decomposing $A(X)$, $C(X)$ and $I(X)$ are supplied to facilitate the discovery of latent factors.
  \item To address confounding variables, three regularization steps are employed: first, $A$ is derived from $X$ with the condition of $A(X) \independent T$ and $A(X)$ being projected onto $Y$; second, $I$ is derived from $X$ with the constraint that $I(X) \independent Y \mid T$, and $I(X)$ should be predictive of $T$; finally, $C(X)$ is balanced across the variation study groups simultaneously.
  \item A distinct regression network is employed for each treatment arm to forecast potential outcomes, with input variables $C(X)$ and $A(X)$ and predicted outcomes $Y^{0}$ and $Y^{1}$.
\end{itemize}
In the process of decomposition, the orthogonal regularizer is utilized in the following way:
\begin{equation}
  \begin{array}{l}
    \mathcal{L}_{O} = \bar{W}^{T}_{I}\cdot \bar{W}_C + \bar{W}^{T}_{C}\cdot \bar{W}_A + \bar{W}^{T}_{A}\cdot \bar{W}_I,
  \end{array}
\end{equation}
{in order to prevent the representation network from rejecting the inputs, $W$ represents the weight matrix of the neural network's representation space, the maximum value of one is imposed on $\bar{W}_I$, $\bar{W}_C$ and $\bar{W}_A$.} Furthermore, when a hard decomposition is in place, the orthogonal regularizer ensures that each variable in $X$ is associated with only one representation network.

Based on DeR-CFR, CATE's prediction performance has been assessed using CF-CV \cite{CF-CV}, which selects the best model or hyperparameters from potential candidates. A meta-learning approach was combined together with deep networks, theoretical reasoning, and the optimal counterfactual information by Curth, A et al. \cite{CATENets}.

Causative treatment effect estimation has always been concerned with rationally using confounding variables. The decoupling of covariates to learn related confounding variables might help remove selection bias and generate unbiased output estimates. Despite its theoretical nature, this kind of methods has also some limitations in practical applications, as it requires decomposing covariates into reasonable explanations.

\subsection{GANs-based counterfactual simulation}
In deep generative models, generative adversarial networks might capture the uncertainty of counterfactual distributions. The generator produces counterfactual results or consistency distribution between control and treatment groups, while the discriminator fits the unbias estimation of the treatment effects.  As well as using factual data, GAN networks also consider the accuracy of counterfactual results when making causal effect estimations. In light of this, generative adversarial models are increasingly used for causal effect estimation.

\begin{figure}[H]
    \centering
    \includegraphics[width=0.78\columnwidth]{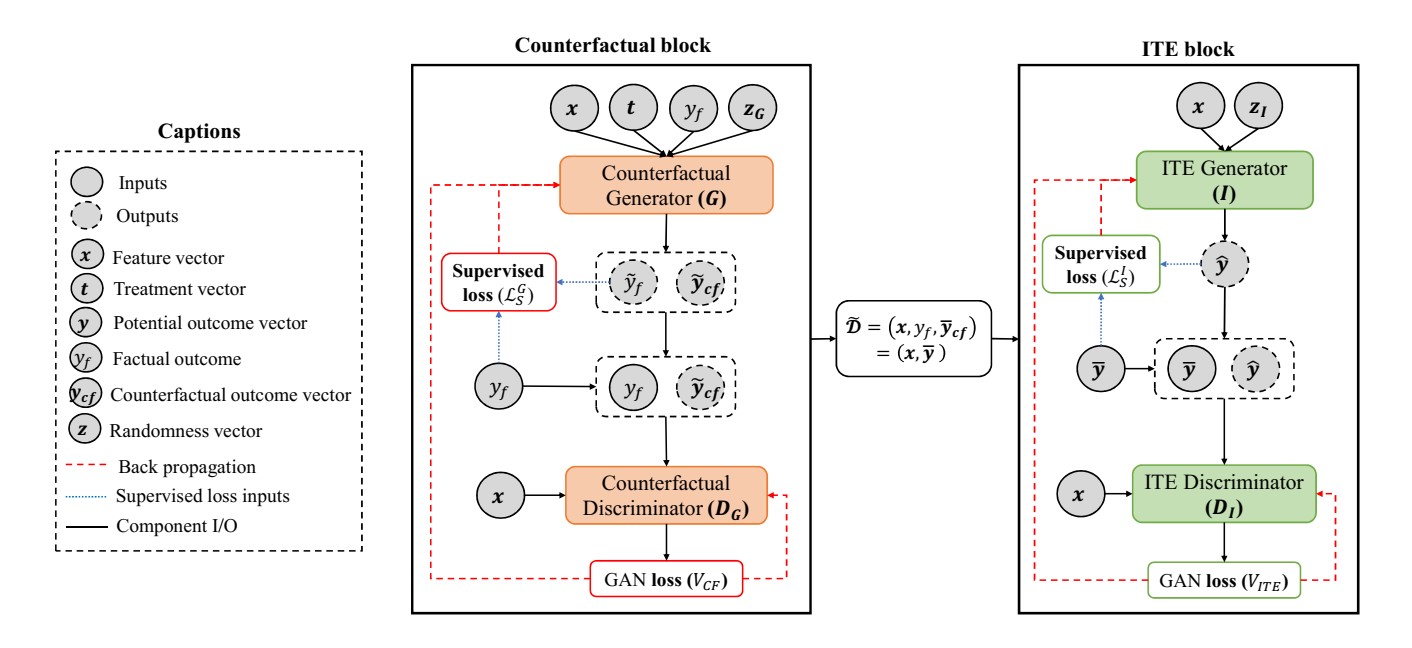}
    \caption{\textbf{Generative Adversarial Nets for inference of Individualized Treatment Effects (GANITE)}: Block diagram \cite{GANITE}}
    \label{fig:GANITE}
\end{figure}

The first approach proposed by Yoon et al. is for the GANITE \cite{GANITE} network to generate the counterfactual results based on factual data and pass them to the ITE generator. Figure \ref{fig:GANITE} shows the detailed block diagram of GANITE.

For a given feature vector $x$, GANITE generates the potential output results by first generating factual outputs $y_f$ and then counterfactual samples $\widetilde{y}_{cf}$ using generator $G$. After these counterfactual data are combined with the original data, a complete data set $\widetilde{D}$ is generated in generator $I$ of the ITE module, which then optimizes $\widetilde{D}$, yielding an unbiased estimate of each treatment's effect.

For the first time, CEGAN \cite{CEGAN} has applied the GAN network to balance the distributions between the spatial treatment group and the control group by leveraging  the discriminative loss of the GAN network and weighting the Decoder's construct loss or weight after the Encoder. In order to solve the generator-discriminator min-max problem, the following reward function is used: 
\begin{equation}
\min _{\left(\theta_{E}, \theta_{I}, \theta_{P}\right)} \max _{\theta_{D}} \mathbb{E}_{q_{E}(\mathbf{z}, \mathbf{x}, t, \mathbf{y})}[\log (D(\hat{\mathbf{z}}, \mathbf{x}, t, \mathbf{y}))]+\mathbb{E}_{q_{P}(\mathbf{z}, \mathbf{x}, t, \mathbf{y})}[\log (1-D(\mathbf{z}, \mathbf{x}, t, \hat{\mathbf{y}}))],
\end{equation}
these estimates might be expressed as $(D(\hat{\mathbf{z}}, \mathbf{x}, t, \mathbf{y}))$ and $(1-D(\mathbf{z}, \mathbf{x}, t, \hat{\mathbf{y}}))$, respectively, as the encoder-decoder's joint distribution, ${q_{E}(\mathbf{z}, \mathbf{x}, t, \mathbf{y})}$ and ${q_{p}(\mathbf{z}, \mathbf{x}, t, \mathbf{y})}$. Different samples belong to variation distributions according to the discriminator.

In addition to GANITE and CEGAN, there are also many  works using GAN networks to estimate the causal effects in other fields. As part of a generative adversarial framework, GAD \cite{GAD} applied GAN networks to continuous treatment problems to learn a sample-balanced weight matrix, which removes the association between treatment regimens and covariates; to address multiple treatment as well as consecutive doses treatment problems, DRGAN \cite{DRGAN} proposed a model architecture consisting of a couterfactual generator, discriminator, and inference block; as a means of better coping with continuous intervention problems, SCIGAN \cite{SCIGAN} added a hierarchical discriminator based on DRGAN. In addition, CTAM \cite{CTAM} has also applied the generative adversarial ideas to the treatment effect estimation of text sequence information. It filters out information related to approximate instrumental variables when learning representations, and matches between the learned representations; to eliminate the association between the treatment and patient history, CRN \cite{CRN} utilized a counterfactual recurrent neural network to reflect the time-varying treatment effect; in ABCEI \cite{ABCEI}, the covariate distributions between the control and treatment groups are well balanced with GAN networks, and a regularization function of mutual information estimators is added to reduce bias; to learn the balanced covariate representations, CETransformer \cite{CETransformer} combined the attention mechanism with WGAN; in TransTEE \cite{TransTEE}, Transformer was used for covariate representation, in which the treatment effectiveness is estimated by the Propensity Score Network and the selection bias might be overcome by the GAN network. In particular, the model might also be used for discrete, continuous, structured or dose-related treatments.

In general, it is not hard to extend the ways of individual treatment effect estimation to multiple interventions and continuous interventions using the GAN network, and it has a significant effect on the balance of representation distribution and the generation of potential results. However, due to lacking a complete theoretical support system, using GAN networks to solve the problem of causal effect estimation requires more impeccable theoretical derivation in the future.

\subsection{Time series causal estimation}

In treatment effect estimation, most models focus on numerical variables, and it is still difficult to deal with textual information and time series information \cite{CITS}. Variable decoupling for textual information estimation might reduce estimation bias since there are many covariates in textual information that are unrelated to causal effect estimation. When dealing with time-series information, RNNs have familiarly been combined to create counterfactual recurrent networks based on historical information.

\begin{figure}[H]
    \centering
    \includegraphics[width=0.84\columnwidth]{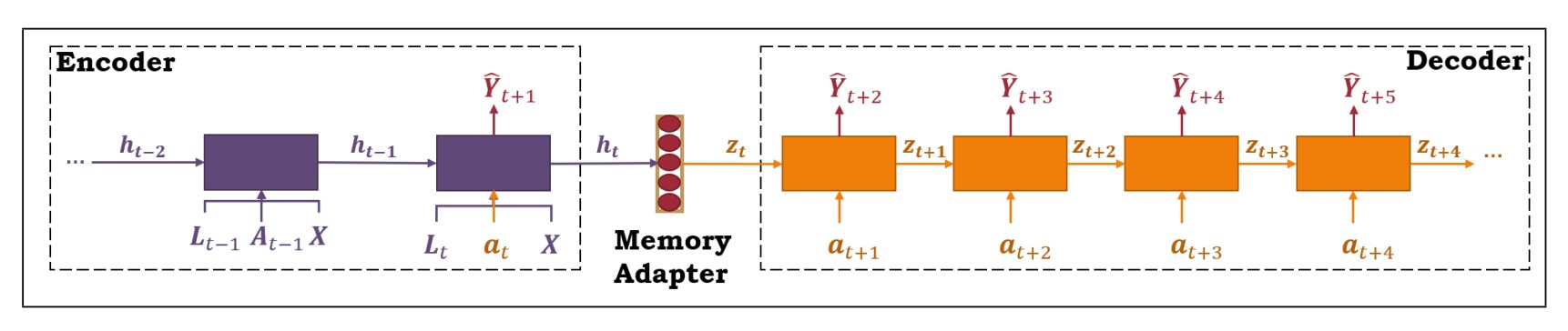}
    \caption{\textbf{Recurrent Marginal Structural Network (R-MSN)}: Architecture for multi-step treatment response prediction \cite{R-MSN}}
    \label{fig:R-MSN}
\end{figure}

The R-MSN \cite{R-MSN} model is first proposed by Lim et al. in order to address the problems arising with continuous treatment and multi-treatments under time series. Figure \ref{fig:R-MSN} illustrates the model's frame structure, which uses a recurrent edge network to remove time-dependent confounding, an a standard RNN structure to encode and decode.

To predict the causal effect, R-MSN used the standard LSTM structure, dividing multi-treatment and continuous intervention problems according to the corresponding time interval. As a counterfactual recurrent network, CRN \cite{CRN} constructs a treatment-invariant representation for each time step based on R-MSN \cite{R-MSN}, eliminating the patient's medical history association between treatment allocation and treatment allocation and balances time-varying confounding biases; by combining with the current treatment assignment and historical information, the hidden confounders are inferred using recurrent weighted neural networks in DSW \cite{DSW}, an then reweighted using time-varying inverse probabilities. In addition to building a multi-task output RNN factor model, TSD \cite{TSD} allocates multiple treatment over time, an then estimates the treatment effects with multi-cause hidden confounders by which the latent variables free of treatment might be inferred. Furthermore, it substitutes the unobserved confounders with latent variables, and infers logistic regression results in the absence of treatment; in SyncTwin \cite{SyncTwin}, treatment estimation is performed based on the temporal structure of the prediction results, and synthetic twin samples are constructed and counterfactual predictions are obtained. 

\begin{figure}[H]
    \centering
    \includegraphics[width=0.62\columnwidth]{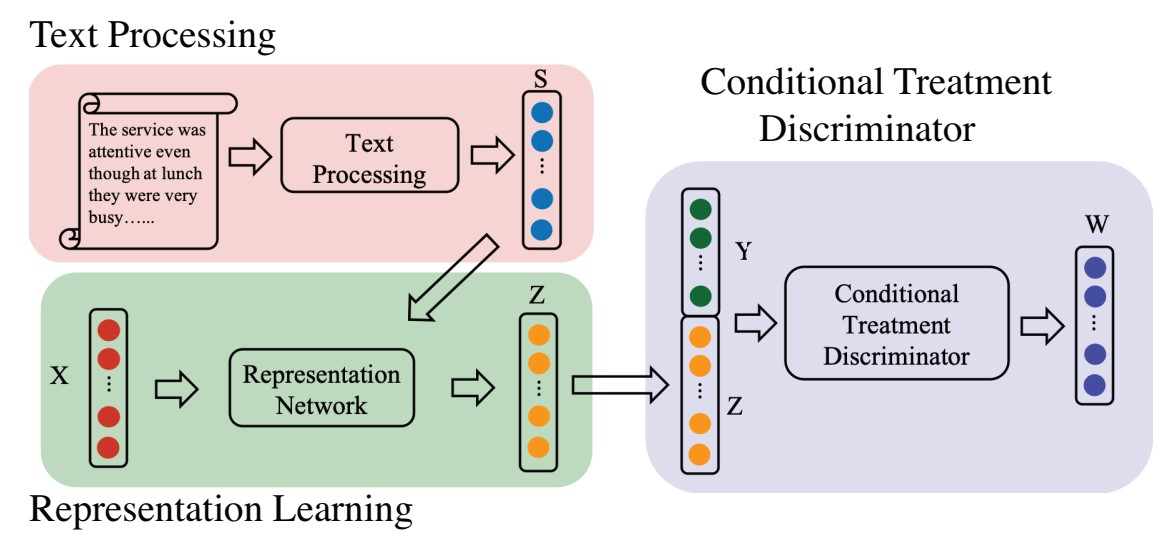}
    \caption{\textbf{Conditional Treatment-Adversarial learning based Matching Method (CTAM)}: Framework \cite{CTAM}} 
    \label{fig:CTAM}
\end{figure}

Yao et al. proposed a matching treatment-adversarial learning CTAM \cite{CTAM} method that takes into account text sequence information. As shown in Figure \ref{fig:CTAM}, when learning representations, it filters out  the approximate instrumental variables and then makes matching between the learned representations to estimate the treatment effect. 

In particular, CTAM \cite{CTAM} consists of three parts: text processing, representation learning, and conditional treatment discrimination. As the first step, the text processing part converts the original text into a vector representation $S$, adds non-text covariates $X$, and then transforms the output into a latent representation $Z$. As a next step, an arithmetic unit between the representation learning network and the conditioned treatment discriminator is calculated during the training process between $Z$ and $Y$. In order to filter out instrumental variables, the representation learning network prevents treatment assignment to correlative variables. As a last step, it implements the match in the representation space $Z$.

To predict treatment assignment using mutual information between global feature representations and individual feature representations, IDRL \cite{IDRL} propsoed to learn Infomax and domain-independent representations. To maximize the ability of capturing  the common prediction information among treatment and control groups, the influence of instrumental variables and irrelevant variables were filtered out. In addition, the SCRNet \cite{SCRNet} attempted to estimate ITE with variation types of variables by dividing the the covariates. Recently, CT \cite{CT} adopted transformer and lstm to capture complex long-term dependencies between time-varying confounders and proposed a new counterfactual domain confusion loss to address the confusion bias.

Estimation of causal effects of textual time series is often combined with the problems related to multiprocessing and continuous processing. Despite the widespread application of this direction, researchers need to develop a standard for measuring intervention effects based on the actual situation, and it is difficult to assess the rationality and reliability of the various working standards used in the industry.

\subsection{Multi-treatment and continuous models}
Casual estimation for individual treatments focuses on solving binary treatment problems, and extending it to multiple treatment is computationally expensive. However, multiple treatment and continuous treatment models have many applications, such as radiotherapy, chemotherapy and surgery for cancer treatment, as well as prolonged usage of vasopressors for many years. It is therefore beneficial to estimating the effects of ongoing interventions in these various treatment settings in order to make significant long-term process decisions.

For the first time, Schwab et al. have extended individual treatment estimation to multi-discrete treatment problems with the PM \cite{PM} algorithm. In PM , counterfactual reasoning is utilized in small batches by matching nearest neighbors samples, which makes it easily implemented and is compatible with a wide range of architectures, and there is no need to increase computation complexity or other hyperparameters for treating arbitrary quantity of patients. In order to capture higher-order effects, TECE-VAE \cite{TECE-VAE} modeled the dependence between treatments by using task embedding, extending the problem to arbitrary subsets of multi-treatment situations. 

When solving problems involving multi-treatments and continuous treatments, GAN networks are frequently combined. A two-step generative adversarial de-aliasing algorithm proposed by GAD \cite{GAD} might be used for continuous treatment problems, removing the association between covariates and treatment variables. Specifically, it is along with the following three steps: A) Creating uncorrelated covariates helps eliminate potential correlations that may introduce bias into the data. B) Addressing sampling bias in the observed data is achieved by assigning sample weights that accurately represent the underlying population. C) Using generative adversarial networks to create synthetic data that is free from biases might improve the quality of the data, leading to more accurate analysis and modeling.

As an improved GAN model, DRGAN \cite{DRGAN} takes the form of a generator, a discriminator, and prediction network to generate a complete dose-response curve for each sample by considering both multi-treatment and continuous treatment options; by using a hierarchical discriminator on the basis of DRGAN, SCIGAN \cite{SCIGAN} was proposed to improve the model's ability of handling the problems of continuous intervention. 

Meanwhile, a set of open model benchmark parameters, including MISE, DPE, PE, and model selection criteria, were proposed in DRNet \cite{DRNet}, which allows the generation of dose-response curves for an unlimited number of treatments under continuous parameters. In VCNet \cite{VCNet} that utilizes a variable coefficient neural network, a continuous ADRF \cite{ARDF_1,ARDF_2,ARDF_3} estimator is automatically calculated for the continuous activation function, which is helpful of preventing the processing information from being lost. Moreover, the existing target regularization method was also extended to obtain a double robust ADRF curve estimator. 

\begin{figure}[H]
    \centering
    \includegraphics[width=0.82\columnwidth]{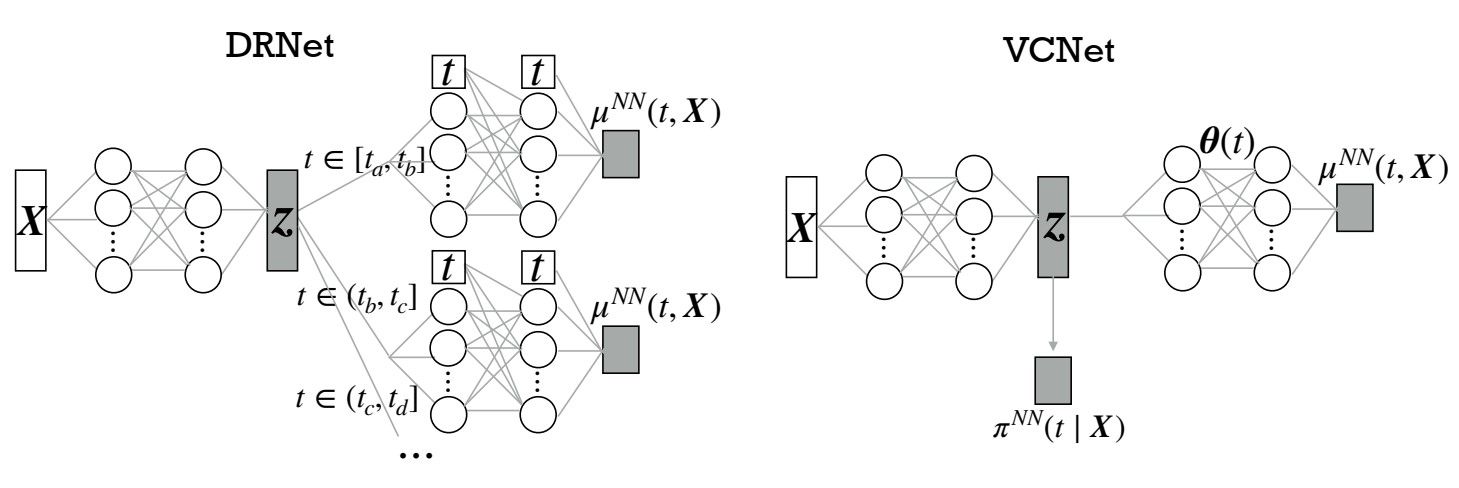}
    \caption{\textbf{Dose Response Network (DRNet) and Varying Coefficient Neural Network (VCNet)}: Comparison of network structures \cite{VCNet}}
    \label{fig:DRNet}
\end{figure}

As part of DRNet \cite{DRNet}, continuous treatments are divided into blocks and trained separately into hidden layers, which are then nested into each other to construct a piecewise fit of individual dose-response curves; a continuous prediction head of weighted treatment is created by VCNet \cite{VCNet} by paying closer attention to treatment continuity, and optimizing the individual prediction head into a mapping function of covariates that change with treatment. The structure comparison of DRNet and VCNet models is shown in Figure \ref{fig:DRNet}. 

Building upon DRnet and VCnet, ADMIT \cite{ADMIT} employed a re-weighting scheme and the IPM to estimate an upper bound on the counterfactual loss and obtain the ADRF. Both theoretical and experimental evidence supported the approach. TransTEE \cite{TransTEE} combined the hierarchical discriminator of SCIGAN \cite{SCIGAN} with the variable coefficient structure of VCNet \cite{VCNet} and inducted the Transformer multi-headed attention mechanism generic framework to extend the causal effect estimation problem to discrete, continuous, structured, and dose-related treatments. Specifically, in MetaITE \cite{MetaITE}, the notion of meta-learning was introduced to the problem of estimating the effects of multiple therapies. A meta-learner was trained from a set of source treatment groups with sufficient samples, and the model was updated through gradient descent with limited samples in the target treatment group.

As the first study of the multi-treatment combination problem, NCoRE \cite{NCoRE} has used the cross-treatment interaction to infer the causal generative processes underlying multiple treatment combinations, in which the counterfactual representations learned in a treatment setting are combined.

To estimate the multi-cause perturbation treatment effect, Prichard et.al  proposed the idea of SCP \cite{SCP} for the first time. To overcome the confounding bias in two steps, a single-cause CATE estimator is first applied to augment observed data and estimate potential outcomes; as a next step, the augmented data set is adapted for covariates to obtain multi-factor unbiased estimators. In addition to illustrating the relationship between single-factor and multiple-factor problems, SCP shows the equivalence of conditional expectations of single-factor interventions and multiple-factor interventions, which might be verified by the following equation:

\begin{equation}
\mathbb{E}\left(Y\left(a_{k}, \mathbf{a}_{-k}\right) \mid \mathbf{X}\right)=\mathbb{E}\left(Y\left(a_{k}\right) \mid \mathbf{X}, \mathbf{A}_{-k}\left(a_{k}\right)=\mathbf{a}_{-k}\right),
\label{Equ:SCP}
\end{equation}
{two sides of the equation represent the equivalence distributions of the conditional expectations for single-cause and multi-cause PO, respectively. To expand the dataset, the initial step involves adding the outcome variable $Y\left(a_{k}\right)$ and the observation variable $\mathbf{A}_{-k}\left(a_{k}\right)$. Employing a supervised learning model on this expanded dataset permits the estimation of the expected value on the right side of \ref{Equ:SCP}, alongside the multifactorial intervention effect on the left, thereby bolstering the estimator's generality.} In contrast to SCP, OOSR \cite{OOSR} proposed a prediction model using a reweighting way that puts emphasize on the outcome-oriented treatment.

Recently, more and more researchers have taken interest in the problem of multi-treatment and continuous treatment, and have also made significant contributions. Nevertheless, there are still many models in this area that need to be developed. Especially, it is still an urgent issue to solve how to formulate a unified causal effect measurement standard.

\section{Current Status of Industrial Applications}
\label{sec:Application}
%Deep learning-based causal effect estimation has been widely applied to data-driven decision making across domains, and plays an important role in incentive allocation, debiased recommendation, precision medicine, and identifying and measuring strategic decision-making. In this section, typical applications in related fields are introduced according to their corresponding data characteristics, we will focus on industrial applications using deep causal models, involving marketing, E-commerce, finance, medicine, economics, and education, and cast concerns on their social value and application value.

Deep causal models have been widely applied to data-driven decision making across domains. In this section, we select six application scenarios and introduce their applications, including marketing, e-commerce, finance, economics, medicine, and education. The reason for choosing these application scenarios is that they are important in real life, and causal inference might bring substantial improvements to these fields. They cover various aspects of human activities and socio-economic systems, and causal inference might help reveal the causal relationships within them and provide decision-makers with more reliable information and guidance. 

Marketing and e-commerce are important components of the modern business field. Causal inference might help companies understand the impact of variation marketing strategies, advertising placements, and promotional activities on sales and consumer behavior. In e-commerce, causal inference might help optimize recommendation systems, personalized marketing, and user experience, enhancing the efficiency and competitiveness of e-commerce platforms.
The financial and economic fields involve investment decisions, market forecasting, economic policies, and more. Causal inference might help analyze the impact of variation factors (such as interest rates, inflation rates, policy measures) on financial markets and economic development, as well as the effectiveness of variation investment strategies and economic policies. This helps financial institutions and policymakers make wiser and more effective decisions.
The medical field involves disease treatment, drug development, clinical trials, and more. Causal inference might help analyze the impact of treatment methods, drugs, and disease prevention measures on patients' health outcomes, as well as the causal relationships between patient characteristics and treatment effects. This contributes to improving the scientificity of medical decision-making and the accuracy of personalized treatment.
The educational field involves student academic performance, teaching methods, and evaluation of educational policies. Causal inference might help educational researchers analyze the impact of variation educational measures, teaching methods, and school environments on students' academic development and educational outcomes. This helps optimize educational policies, improve teaching practices, and enhance student learning performance.

Deep causal models have advantages over traditional methods in representation learning, handling high-dimensional data, and adapting to complex structures. In terms of representation learning, traditional methods often require feature selection and extraction, which might be challenging for complex data and problems. Deep learning models, on the other hand, might automatically learn the feature representations of data without the need for feature engineering. This allows the model to adapt to variation types of data and better capture the causal relationships hidden in the data. Traditional causal inference methods typically use linear models to model causal relationships, but real-world causal relationships are often nonlinear. Deep learning models have powerful nonlinear modeling capabilities and might learn complex nonlinear causal relationships, leading to more accurate causal inference. Traditional methods may be limited in handling high-dimensional data, while deep learning models have the ability to handle high-dimensional data. Deep learning models might process datasets with a large number of features and perform more accurate causal inference. Deep learning models might also adapt to various complex structured data, such as graph data and sequence data. Traditional methods may require additional manual feature engineering and model design when dealing with such complex structured data, whereas deep learning models might directly learn complex causal relationships from raw data.

{In traditional causal inference methods, it is common to employ linear models based on the assumption of linear relationships between variables. In contrast, deep learning methods do not make specific assumptions about the distribution of data or the form of relationships between variables, as they possess powerful capabilities for nonlinear modeling that can unveil complex data structures. Real-world datasets often lack counterfactual information, meaning the potential outcomes of individuals under different treatment conditions cannot be directly observed. This absence restricts the computation of ground truths such as the Average Treatment Effect (ATE) \cite{ATE}, the Average Treatment Effect on the Treated (ATT) \cite{Survey}, and the Precision in Estimation of Heterogeneous Effects (PEHE) \cite{IHDP_2}, which in turn affects the assessment of model performance and the determination of optimal parameters. Nonparametric methods are thus favored because they do not rely on specific assumptions. Nonetheless, when ground truths are known and model validation is feasible, finely-tuned deep learning methods often outperform nonparametric or semiparametric approaches, such as Bayesian Additive Regression Trees (BART) \cite{BART}, which is a recognized accurate and popular Bayesian nonparametric method.}

\subsection{Marketing applications}
Towards the goal of maximizing the return on investment, it is crucial to decide the assignment of user-specific incentive under limited budget. Uber and Didi Chuxing have incentivized drivers as well as passengers with commission and coupons to boost ride-sharing business. Mobile payments, such as Apple Pay, Alipay, WeChat Pay, have designed various mobile marketing campaigns to increase the usage of their applications. Amazon and JD Digits attempted to give out discounts to enhance attraction to customers. Incentive allocation relies on treatment effect estimation models to estimate users’ purchase probability with variation incentives. PCAN \cite{ALPM} was proposed to learn an an unbiased model by leveraging a small set of unbiased data. Specifically, a biased network was built to generate unbiased data representation by controlling the distribution difference to a unbiased network. Offline and online experimental results indicated that this model might alleviate the problem of price-bias and lead significant performance improvement for a real-world marketing campaign. Besides, DESCN \cite{DESCN} was developed to integrate the learning process of the treatment propensity, the true responses, and the treatment effect in a joint way simultaneously. It has been validated in voucher distribution business in Lazada, one of the largest E-commerce operators belonging to the Alibaba Group. Related results indicated that this method is advantageous in both performances of ITE estimation and uplift ranking. For multi-treatment scenarios, Amazon presented MEMENTO  \cite{MEMENTO}. Different treatment types all gained matching representation by minimizing the upper bound of the loss, which is the sum of the factual and counterfactual estimation errors. Besides incentive allocation, deep causal models are also applied to other tasks in marketing. For example, a user retention model UR-IPW \cite{COST} was proposed by AntGroup to account for impression-revisit effect, where users might have explicit and implicit interactions with the recommender system. The model made full use of both kinds of interactions from the observed data, and the selection bias caused by user’s self-selection was eliminated by inverse propensity weighting.

\textbf{Practical Application}: In bilateral business relations, effective marketing strategy should also motivate merchants as well as customers. For example, to encourage mobile payment activities, merchants get shared incentives with customers after them scanning QR-codes and paying with Alipay. However, making independent optimization for both sides might be nonoptimal. Therefore, several work  \cite{Graph}  \cite{Joint} took the mutual influence into account, and used graph neural networks to represent merchants and customers jointly by modeling the underlying bipartite influences. Extensive experimental results demonstrate the effectiveness of the proposed approaches.

\subsection{E-commerce applications}
Recommender systems play an important role in E-commerce. Traditional recommender systems learn user preference by mining correlation in observational data, resulting in biases including selection bias, exposure bias, position bias, conformity bias, etc. To address this issue, Amazon, Netflix, Criteo, Alibaba, AntGroup, JD, Kuaishou, etc. have all begun to utilize causal effect estimation to extract causality. 
% Datasets and recent works on recommendation debiasing might refer to  \cite{User} and  \cite{Bias}. 
Debiased information bottleneck was proposed by Huawei  \cite{Survey_7}, which is applicable to various types of bias. The model learnt a biased representation in the training phase, which include a biased part and an unbiased part, and only the unbiased part of representation is used in the test phase to reach more accurate recommendations. Wei et al. \cite{Mitigating} studied the issue of popularity bias from a cause-effect perspective , where multi-task learning is leveraged to model the contribution of each cause, and counterfactual inference is utilized to eliminate the effect of item popularity during testing. Following that, ESCM$^2$ \cite{ESCM} further exploit the sequential pattern of user actions to address the data sparsity issue and employed a counterfactual risk minimizer to simultaneously address the issues of estimation bias in CVR estimation and potential independence priority in CTCVR estimation. Furthermore, in some scenarios, another set of uniform data might also be used to improve model performances by reducing the data bias. Criteo proposed a multi-task objective that jointly factorizes the matrix of biased data and the matrix of uniform data  \cite{recommender_6}. A general knowledge distillation framework that enables uniform data modeling was proposed by Liu, D. \cite{General}, which performs distillation from the view of label, feature, sample, and model structure. Differently, AutoDebias  \cite{Autodebias} employed the technique of bi-level optimization to optimize the debiasing parameters with a small set of uniform data.

\textbf{Practical Application}: CausalMTA  \cite{ESCM} was proposed by Alibaba, which might remove the confounding bias for both static and dynamic perspectives. It is achieved by sample reweighting, which enables learning an unbiased conversion prediction model. Its effectiveness is demonstrated by a real advertisement impression data.

\subsection{{Economic and Financial applications}}
Causal effect estimation has attracted more and more attention of researchers in economic science, especially how to better combine with deep learning to solve practical problems. Learning about cause and effect is arguably the main goal in applied econometrics. Angrist and Steve Pischke coined the term “credibility revolution”  \cite{credibility}. They argued that economics turns to transparent empirical strategies applied to specific causal problems. Problem-driven methodological agendas are mostly based on the propensity score theorem of Rosenbaum and Rubin  \cite{central}. This theorem changes applied econometrics to focus our attention on the process of determining treatment assignment rather than models for outcomes. Dehejia and Wahba  \cite{ATE} was the first to demonstrate the value of this approach. More recently, Belloni et al.  \cite{belloni} utilized deep learning to model scores while modeling outcomes. This work might be seen as an extension of Rubins's notion of dual robustness to a broader class of empirical strategies. Angrist  \cite{2022empirical} introduced a novel framework, the local average treatment effects framework for causal effect estimation, to help make the empirical strategies in economics should be transparent and credible. The LATE theorem might tell us for whom particular instrumental variables (a tool used when an explanatory variable is correlated with the error term in a regression model, due to omitted variable bias, measurement error, or simultaneous causality) and regression discontinuity estimates are valid. Hal R. Varian  \cite{varian} argued that the powerful techniques used in deep learning may be useful for developing better estimates of the counterfactual, potentially improving causal effect estimation. For generalized neighbor matching to estimate individual and average treatment effects, Vikas  \cite{2018deep} proposed the use of deep learning techniques in econometrics, specifically for causal effect estimation and for estimating individual as well as average treatment effects. Recently, Athey and Wager  \cite{efficient} brought in insights from semiparametric efficiency theory in econometrics to propose a new estimator for optimal policies and to analyze the properties of this estimator. Policies might be compared in terms of their “risk”, which is defined as the gap between the expected outcomes using the (unknown) optimal policy and the estimated policy. Arpino et al.  \cite{assessing} proposed an explicitly model, which was applied to the evaluation of a policy implemented in Tuscany on small handicraft firms, to address the violates the Stable Cell Handling Value Assumption due to the interference between cells. Results show that the benefits from the policy are reduced when treated firms are subject to high levels of interference. Additional application examples include rethinking the benefits of youth employment programs by analyzing the heterogeneous effects of summer jobs \cite{economics-app1}; using causal machine learning to estimate the heterogeneity of policy impacts, such as in a case study of health insurance reform in Indonesia \cite{economics-app2}; and employing causal machine learning to evaluate whether training unemployed immigrants should be prioritized in Belgium \cite{economics-app3}.

\textbf{Practical Application}: In the field of Fin Tech, MYbank of Ant Group adopted causal counterfactual inference debiasing methods  \cite{dags,VSR,SCP,NCoRE,regret,assaad} to solve the biased problem of loan marketing AB experiment. In order to effectively measure the intervention effect, causal counterfactual inference can be employed to construct a homogeneous control group from the full experimental population based on observational data, so that horizontal comparisons might be made.

\subsection{Medical applications}
In the area of precision medicine, the research community has started to explore quantitative individual-level effect of a treatment, by using observational data from electronic health records. Clinical decision-makers target at determining the personalized optimal treatment process, using static or time-series observational data. However, observation datasets familiarly face the problem of treatment assignment bias. To address it, a series of models have been proposed. In many medical applications, multiple variables might be intervened at the same time. In order to estimate impact of multi-cause treatment, Single-cause Perturbation \cite{SCP} was proposed, which firstly augments the observational dataset with potential outcomes estimated from single-cause interventions, and then adjusts the covariate of the augmented dataset to learn the estimator. Following that, GraphITE \cite{graphite} was proposed, which utilizes graph neural networks to learn the representation of the graph-structured treatments. In order to reduce the observation biases, HSIC regularization is employed to obtain an independent representation of the targets in regard to the treatments. Observational data also contain information on complex time-varying treatment. To estimate the effectiveness of treatment over time, SCIGAN \cite{SCIGAN} was proposed, which is flexible and has the ablity to estimate counterfactual outcomes for sequential interventions. Additionally, TE-CDE \cite{continuous} was proposed, which might estimate the potential outcomes at any time point. To do so, it uses adversarial training to adjust for time-dependent confounding.

\textbf{Practical Application}: The van der Schaar Lab has a wide range of clinical applications on causality models, including COVID-19, organ transplantation, etc. Particularly, to decide “one best model” for each scenario, a validation procedure is introduced to estimate the performances of variation causal effect estimation methods by using influence functions \cite{validating}.

\subsection{Educational applications}
Education plays a central role in modern society, especially in earlier years  \cite{2005causal}, a wave of new studies on the effects of educational interventions on student performance has emerged. Therefore, the need for evidence-based policy in the field of education is increasingly recognized. Education policy-makers and practitioners want to know which policies and practices might best achieve their goals. However, providing empirical evidence suitable for guiding policy is not an easy task, because it refers to causal effect estimations that require special research methods which are not always easy to communicate due to their technical complexity. From a Bayesian perspective, David  \cite{card,2011econometric} introduced a review and synthesis of the problem of causal inference in large scale educational assessments which requires the articulation of framework for causal effect estimation followed by a statistical approach that closely matches the framework and might yield the causal estimate of interest. Zhao et al.  \cite{edu} proposed the Residual Counterfactual Networks in an Intelligent Tutor System might decide which hint is more suitable for a specific student. However, the effectiveness of an intervention is necessarily multifaceted and complex effects differ between students, as a function of implementation  \cite{sales2016}, and, potentially, as a function of time and location. Sales et al.  \cite{sales2021} explored a variation sort of treatment effect heterogeneity differences in effectiveness for variation outcomes. Specifically, variation posttest items measuring variation skills. Carvalho  \cite{applying} conductd causal effect estimation on students’ online behavior patterns using a variation toolkit, GeNIe3, using data from a learning management system. Chen et al.  \cite{2021affect} developed a causal discovery framework that utilized TETRAD  \cite{tetrad}.

\subsection{{Other applications}}
In addition to the aforementioned applications, causal inference might also be applied in a wide range of fields. (1)Social policy evaluation: Causal inference might be used to evaluate the effectiveness of social policies, such as employment policies, welfare policies, housing policies, etc. By analyzing causal relationships, policymakers might understand the impact of policies, enabling them to make improvements and decisions. (2)Criminology and legal policy: Causal inference might help study the causes and effects of crime and provide a basis for effective legal policies. For example, by analyzing the impact of specific factors on crime rates, more effective crime prevention measures might be developed. (3)Environmental impact assessment: Causal inference might be used to assess the impact of specific activities or projects on the environment, such as the impact of construction projects on ecosystems or the impact of pollution sources on environmental quality. Through causal inference, environmental impacts might be quantified and evaluated, guiding environmental protection and sustainable development decisions. (4)Public health and epidemiology: Causal inference might be applied in public health and epidemiological research, such as studying the impact of vaccination on disease transmission or the impact of food safety measures on food poisoning. Through causal inference, more effective public health policies and measures might be formulated. Causal inference provides a powerful tool and method for scientific research, promoting the development and innovation of disciplines, helping decision-makers understand the contribution of variation factors to outcomes, thereby optimizing resource allocation and making wiser decisions.

\section{Guideline for Experiment}
\label{sec:Experiment}
After a detailed exposition of the profound causal modeling methodology, this section will furnish an elaborate experimental manual, encompassing an exhaustive synopsis and scrutiny of datasets, source codes, and empirical trials.
\subsection{Datasets}
As the observation of counterfactual outcomes is often unattainable in actual situations, the task of identifying datasets that fulfill the experimental prerequisites becomes challenging. Many datasets commonly employed in academic studies are predominantly semi-synthetic in nature to address this limitation. Following the survey \cite{Survey}, we further supplemented in Table \ref{tab:Dataset_Feature} more relevant datasets that are popularly used for causal analysis. Moreover, the utility of these datasets can be inferred by considering their potential application scenarios. In addition, Table \ref{tab:dataset} summarizes the web links to these datasets and the classical models that are evaluated on them. Below are the detailed descriptions of the available datasets.

\textbf{IHDP}.
The Infant Health and Development \cite{IHDP} dataset originated from a controlled trial that randomly assigned preterm infants with low birth weight to either a treatment or a control group. The study collected pre-treatment covariates such as birth weight, head circumference, neonatal health index, and maternal characteristics like age, education, drug, and alcohol use. The treatment group received intensive and high-quality childcare, including specialized home visits \cite{IHDP_2}, while the control group received standard care. The primary outcome measure was the cognitive test score of the infants. To develop unbiased selection models, the study excluded noisy subsets of the treatment group from the analysis.

\textbf{Jobs}.
In the Jobs \cite{Jobs} in LaLonde (1986) study, the employment dataset was utilized to assess the effect of vocational training on employment outcomes. The dataset consisted of both randomized data from state-supported work programs and non-randomized data from observational studies. Pre-treatment covariates, such as age, education, race, and income, were collected for the years 1974 and 1975. The treatment group received vocational training, while the control group did not receive any training. The primary focus was on the employment status of individuals in both groups as the primary outcome variable.

\textbf{Twins}.
The Twins \cite{Twins} dataset consists of data on twin births in the United States between 1989-1991, including 40 pre-treatment covariates related to pregnancy, twin births, and parents. These covariates include gestational weeks, quality of care during pregnancy, pregnancy risk factors such as anemia, alcohol, and smoking, nursing, and residence. The primary outcome measure for each twin pair is the one-year mortality rate. The dataset also features a twin dataset, which categorizes twins into treatment (heavier twin) and control (lighter twin) groups. To simulate selection bias, the study assigns variation treatments based on user-defined criteria.

\textbf{News}.
The News dataset is a sample of 5000 news articles randomly selected from the New York Times corpus. The dataset explores media consumers' perceptions of these news items and includes word counts for each article. The primary outcome measure is readers' opinions of the news item, and the treatments investigated are variation devices used to view the news, such as smartphones, tablets, computers, and TVs. The dataset offers valuable insights into how consumers interact with news articles on various devices and might inform media companies' content delivery strategies.

\textbf{ACIC}.
Every year since 2016, the Atlantic Causal Inference Conference has organized a challenge focused on estimating causal effects from various datasets. The latest conference dataset for this challenge is summarized in a publication titled “ACIC” \cite{ACIC}.

\textbf{TCGA}.
The Cancer Genome Atlas (TCGA) \cite{TCGA} is a comprehensive and extensive database of genomic information, containing billions of genetic sequences. It includes data from 9658 individuals who have undergone variation cancer treatments, such as drug therapy, chemotherapy, and surgical procedures, with the aim of assessing the risk of cancer recurrence or development after treatment.

\textbf{PK-PD model of tumor growth}.
The pharmacokinetic-pharmacodynamics (PK-PD) \cite{PK-PD} model is a valuable tool for exploring dose-response relationships and guiding treatment decisions. This model might be particularly useful in the treatment of non-small cell lung cancer, where factors such as the combined effects of chemotherapy and radiotherapy, cellular regeneration post-treatment, patient prognosis, and variations in tumor size at diagnosis must be considered. By analyzing expected treatment responses \cite{pkpd_1,pkpd_2}, the PK-PD model might predict tumor growth in the presence of time-dependent confounding variables. This model is widely used by clinicians to generate hypotheses and identify optimal treatment options.

\textbf{MIMIC III}.
MIMIC III \cite{MIMIC} is an electronic health records database that contains data from ICU patients. The benchmark dataset comprises 7413 samples and 25 covariates after filtering out missing values. In the ICU, common treatment options for sepsis patients include antibiotics, vasopressors, and mechanical ventilators. Vital signs such as white blood cell count, blood pressure, and oxygen saturation are used over time to evaluate the effects of these treatments on patients. For a detailed description of the clinical data, please refer to the article \cite{2011mimic}.

\textbf{NICO}.
The NICO \cite{LaCIM} image dataset used for object classificationn may exhibit bias in sample selection. The “animals” dataset, which is often used for cat or dog classification, is considered a benchmark distribution for non-i.i.d. data. However, this bias may arise due to various factors such as the timing and criteria for selecting images, the context in which animals are presented, and the semantic shape of the animals. Additionally, environmental factors like the presence of “grass” or “snow” might also influence classification results.

\textbf{ADNI}.
The ADNI \cite{ADNI} dataset contains three latent representation outputs: Alzheimer's Disease, Mild Cognitive Impairment, and Normal Control. Covariates in the dataset include age and TAU \cite{TAU}, which might be utilized to determine the suitability of Magnetic Resonance Imaging (MRI) as a therapeutic input.

\textbf{CHESS}.
The CHESS \cite{COVID_1,COVID_2} dataset was designed to collect information about ICU patients during the first wave of the COVID-19 pandemic in England. The dataset includes risk factors, treatments, and outcomes for 3090 patients. Some of the covariates in the dataset are age, multiple morbidity, ventilation, and antiviral drug treatments. The primary outcome measured in the dataset is the length of stay in the ICU \cite{ICU}.

\textbf{CPRD}.
The Clinical Practice Research Datalink (CPRD) \cite{CPRD} is a database containing medical records from NHS general practice clinics in the UK, covering around 6.9 $\%$ of the country's population. It has been found to be associated with secondary care admissions based on national mortality records and hospital event statistics. The study involves measuring low-density lipoprotein levels after CPRD initiation, with treatment defined as the date of the first prescription. Prior to treatment initiation, the following risk factors are measured as temporal covariates: high-density lipoprotein cholesterol, blood pressure, pulse, creatinine, triglycerides, and smoking status. Participants for the HPS registry are selected from a pool of 125,784 eligible individuals. The treatment and control groups are divided into three equally sized subsets for training, validation, and testing, resulting in a total of 17,371 treatment groups and 24,557 control groups.

\textbf{BlogCatalog}.
BlogCatalog \cite{Blogcatelog} is an online community where bloggers might post their content, and social relationships between bloggers are represented as edges in the dataset. Each blogger is represented as an instance, and the blog descriptions are represented using keywords in a bag-of-words format. By analyzing reader opinions, the study investigates whether blog content gets more comments on mobile or desktop devices. The study uses the content of readers' comments on mobile devices versus desktop devices to estimate parameters for individual treatment effects. If a blogger's content is read more on a mobile device than on a desktop device, the blogger belongs to the treatment group, and vice versa.

\textbf{Flickr}.
Flickr \cite{flickr} is an online social networking platform where users might upload and share photos and videos. The dataset comprises instances representing individual users and edges that indicate their social relationships. Each user's features are represented by tags of interest. The study's settings and assumptions are similar to those of the BlogCatalog dataset, with the aim of identifying which tags are associated with increased engagement on the platform. The study also seeks to estimate individual treatment effects using parameters derived from user comments and engagement with photos and videos.

\begin{table*}[t]
\scriptsize
\renewcommand\arraystretch{2}
\renewcommand\tabcolsep{25.0pt} 
\caption{\textbf{Overview of some open available datasets for causal analysis and their application scenario}}
\label{tab:Dataset_Feature}
\begin{center}
\vspace{0.2cm}
\resizebox{\textwidth}{!}{
\begin{threeparttable} 
\begin{tabular}{c@{\hspace{1cm}}c@{\hspace{1cm}}c@{\hspace{1cm}}c@{\hspace{1cm}}c@{\hspace{1cm}}c}
\toprule
 \normalsize \textbf{Datasets} & \normalsize \textbf{Binary} & \normalsize \textbf{Multiple} & \normalsize \textbf{Continuous treatment} & \normalsize \textbf{Times series} & \normalsize \textbf{Structured}\\
 \hline
 \normalsize IHDP&\Checkmark&\Checkmark&&&\\
 \normalsize Jobs&\Checkmark&&&&\\
 \normalsize Twins&\Checkmark&&&&\\
 \normalsize News&\Checkmark&\Checkmark&\Checkmark&&\\
 \normalsize TCGA&&&\Checkmark&&\Checkmark\\
 \normalsize ACIC&&\Checkmark&\Checkmark&&\\
 \normalsize Tumor Growth&&&&\Checkmark&\\
 \normalsize MIMIC III&&&\Checkmark&&\\
 \normalsize NICO&\Checkmark&&&&\\
 \normalsize CMNIST&\Checkmark&&&&\\
 \normalsize ADNI&&\Checkmark&&&\\
 \normalsize CHESS&&\Checkmark&&&\\
 \normalsize CPRD&&&&\Checkmark&\\
 \normalsize BlogCatalog&&\Checkmark&&&\Checkmark\\
 \normalsize Flickr&&\Checkmark&&&\Checkmark\\
 \bottomrule
\end{tabular}
\end{threeparttable}
}
\end{center}
\end{table*}

\begin{table*}[ht]
\normalsize
\renewcommand\arraystretch{2}
\renewcommand\tabcolsep{25.0pt} 
\caption{\textbf{The web links to the open available datasets and the related models that have used them for performance evaluation}} 
\label{tab:dataset}
\begin{center}
\vspace{0.2cm}
\resizebox{\textwidth}{!}{
\begin{threeparttable} 
\begin{tabular}{c@{\hspace{1cm}}c@{\hspace{1cm}}c}
\toprule
 \normalsize \textbf{Datasets} & \normalsize \textbf{Links} & \normalsize \textbf{Methods} \\
 \hline
 \normalsize IHDP& \url{https://www.fredjo.com}& \cite{BNN,DCN-PD,CFRNet,CEVAE,Deep-Treat,RCFR,GANITE,SITE,PM,CTAM,Dragonnet,ACE,RSB,DKLITE,DR-CFR,ABCEI,BWCFR,VCNet,CETransformer,DeR-CFR,FlexTENet,CATENets,CETransformer,TransTEE,SCRNet}\\
 \normalsize Jobs&\url{https://www.fredjo.com}& \cite{CFRNet,CEVAE,GANITE,SITE,ACE,DKLITE,ABCEI,CETransformer,DONUT,SCI,SCRNet,IDRL,NESTER,CITE,CBRE}\\
 \normalsize Twins&\url{www.nber.org/data/linked-birth-infant-death-data-vital-statistics-data.html}& \cite{CEVAE,GANITE,CEGAN,SITE,ACE,DKLITE,GAD,ABCEI,CETransformer,DONUT,NESTER,CITE,CBRE}\\
 \normalsize News&\url{https://archive.ics.uci.edu/ml/datasets/bag+of+words}& \cite{BNN,PM,CTAM,DRGAN,SCIGAN,SCI,DRNet,VCNet,TransTEE,IDRL,MetaITE,ADMIT}\\
 \normalsize ACIC&\url{https://jenniferhill7.wixsite.com/acic-2016/competition}& \cite{Dragonnet,FlexTENet,ABCEI,CRE,BCF-IV,ITRS,CATE,DESCN}\\
\normalsize TCGA&\url{https://gdc.cancer.gov/}& \cite{DRGAN,SCIGAN,SCI,TransTEE,OOSR,ADMIT,TCGA_1,TCGA_2,TCGA_3,TCGA_4,TCGA_5}\\
 \normalsize Tumor Growth&\url{www.nature.com/scientificreports/}& \cite{R-MSN,CRN,TG_1,TG_2,TG_3,TG_4}\\
 \normalsize
 MIMIC III&\url{https://mimic.physionet.org/}& \cite{TSD,ABCEI,SCIGAN,DRNet,SCI,CT,MIMIC_1}\\
 \normalsize
NICO&\url{https://nico.thumedialab.com/}& \cite{LaCIM,NICO_1,NICO_2,NICO_3,NICO_4,NICO_5}\\
 \normalsize
ADNI&\url{https://adni.loni.usc.edu/}& \cite{LaCIM,ADNI_1,ADNI_2,ADNI_3,ADNI_4}\\
 \normalsize
CHESS&\url{https://www.heywhale.com/mw/dataset/5e8ee81fe7ec38002d00f9cb}& \cite{SCP,COV_1,COV_2,COV_3}\\
 \normalsize
CPRD&\url{https://academic.oup.com/ije/article/44/3/827/632531}& \cite{SyncTwin,CPRD_1,CPRD_2,CPRD_3,CPRD_4,CPRD_5}\\
 \normalsize
BlogCatalog&\url{https://www.blogcatalog.com}& \cite{NETDECONF,BlogCatalog_1,BlogCatalog_2}\\
 \normalsize Flickr&\url{https://www.flickr.com}& \cite{NETDECONF}\\
 \bottomrule
\end{tabular}
\end{threeparttable}
}
\end{center}
\end{table*}

\subsection{Source codes}
Most existing reviews are limited to an overview of traditional algorithm code. Here we summarize the source code associated with deep learning and programming frameworks used. In academia and industry, PyTorch and TensorFlow are the mainstream deep learning frameworks, with more than 80 $\%$ scholars and programmers using them for model construction.  In addition to the popularly used datasets as motioned above, we also list in Table \ref{tab:code} the available source codes for some representative models and the relevant dataset they have used.

\begin{table*}[ht]
\scriptsize
\renewcommand\arraystretch{2}
\renewcommand\tabcolsep{25.0pt} 
\caption{\textbf{Overview of some available datasets and web links to them}}
\label{tab:code}
\begin{center}
\vspace{0.2cm}
\resizebox{\textwidth}{!}{
\begin{threeparttable} 
\begin{tabular}{c@{\hspace{1cm}}c@{\hspace{1cm}}c@{\hspace{1cm}}c}
\toprule
\normalsize \textbf{Methods} & \normalsize \textbf{Datasets} & \normalsize \textbf{Framework} & \normalsize \textbf{Links} \\
\hline

 \normalsize DCN-PD \cite{DCN-PD}&\normalsize IHDP&\normalsize Pytorch&\url{https://github.com/Shantanu48114860/Deep-Counterfactual-Networks-with-Propensity-Dropout}\\
 \normalsize BNN \cite{BNN},CFRNet \cite{CFRNet}&\normalsize IHDP,Jobs,News&\normalsize Tensorflow&\url{https://github.com/clinicalml/cfrnet}\\
 \normalsize CEVAE \cite{CEVAE}&\normalsize IHDP,Twins,Jobs&\normalsize Tensorflow&\url{https://github.com/AMLab-Amsterdam/CEVAE}\\
 \normalsize GANITE \cite{GANITE}&\normalsize IHDP,Twins,Jobs&\normalsize Tensorflow&\url{https://github.com/jsyoon0823/GANITE}\\
 \normalsize SITE \cite{SITE}&\normalsize IHDP,Twins,Jobs&\normalsize Tensorflow&\url{https://github.com/Osier-Yi/SITE}\\
 \normalsize R-MSN \cite{R-MSN}&\normalsize PK-PD model of tumor growth&\normalsize Tensorflow&\url{https://github.com/sjblim/rmsn_nips_2018}\\
 \normalsize PM \cite{PM}&\normalsize IHDP,News&\normalsize Tensorflow&\url{https://github.com/d916b/perfect_match}\\
 \normalsize Dragonnet \cite{Dragonnet}&\normalsize IHDP,ACIC&\normalsize Tensorflow&\url{https://github.com/claudiashi57/dragonnet}\\
 \normalsize DKLITE \cite{DKLITE}&\normalsize IHDP,Twins,Jobs&\normalsize Tensorflow&\url{https://github.com/vanderschaarlab/mlforhealthlabpub/tree/main/alg/dklite}\\
 \normalsize CRN \cite{CRN}&\normalsize PK-PD model of tumor growth&\normalsize Tensorflow&\url{https://github.com/vanderschaarlab/mlforhealthlabpub/tree/main/alg/counterfactual_recurrent_network}\\
 \normalsize TSD \cite{TSD}&\normalsize MIMIC III&\normalsize Tensorflow&\url{https://github.com/vanderschaarlab/mlforhealthlabpub/tree/main/alg/time_series_deconfounder}\\
 \normalsize ABCEI \cite{ABCEI}&\normalsize IHDP,Twins,Jobs,ACIC,MIMIC III&\normalsize Tensorflow&\url{https://github.com/octeufer/Adversarial-Balancing-based-representation-learning-for-Causal-Effect-Inference}\\
 \normalsize LaCIM \cite{LaCIM}&\normalsize NICO,CMNIST,ADNI&\normalsize Pytorch&\url{https://github.com/wubotong/LaCIM}\\
 \normalsize SCIGAN \cite{SCIGAN}&\normalsize TCGA,News,MIMIC III&\normalsize Tensorflow&\url{https://github.com/ioanabica/SCIGAN}\\
 \normalsize DRNet \cite{DRNet}&\normalsize TCGA,News,MIMIC III&\normalsize Tensorflow&\url{https://github.com/d909b/drnet}\\
 \normalsize VCNet \cite{VCNet}&\normalsize IHDP,News&\normalsize Pytorch&\url{https://github.com/lushleaf/varying-coefficient-net-with-functional-tr}\\
 \normalsize DeR-CFR \cite{DeR2022}&\normalsize IHDP,Jobs,Twins&\normalsize Tensorflow&\url{https://github.com/anpwu/DeR-CFR}\\
 \normalsize DONUT \cite{DONUT}&\normalsize IHDP,Twins,Jobs&\normalsize Tensorflow&\url{https://github.com/tobhatt/donut}\\
 \normalsize FlexTENet \cite{FlexTENet},CATENets \cite{CATENets}&\normalsize IHDP,Twins,ACIC&\normalsize Jax,Pytorch&\url{https://github.com/AliciaCurth/CATENets}\\
 \normalsize SCP \cite{SCP}&\normalsize COVID-19&\normalsize Pytorch&\url{https://github.com/vanderschaarlab/Single-Cause-Perturbation-NeurIPS-2021}\\
 \normalsize SyncTwin \cite{SyncTwin}&\normalsize CPRD&\normalsize Pytorch&\url{https://github.com/vanderschaarlab/SyncTwin-NeurIPS-2021}\\
 \normalsize TransTEE \cite{TransTEE}&\normalsize IHDP,News,TCGA&\normalsize Pytorch&\url{https://github.com/hlzhang109/TransTEE}\\
 \normalsize CF-CV \cite{CF-CV}&\normalsize IHDP&\normalsize Tensorflow&\url{https://github.com/usaito/counterfactual-cv}\\
 \normalsize CGN \cite{CGN}&\normalsize MNIST,ImageNet&\normalsize Pytorch&\url{https://github.com/autonomousvision/counterfactual_generative_networks}\\
 \normalsize DESCN \cite{DESCN}&\normalsize ACIC,Epilspsy&\normalsize Pytorch&\url{https://github.com/kailiang-zhong/DESCN}\\
 \normalsize CT \cite{CT}\normalsize &MIMIC III&\normalsize Pytorch&\url{https://github.com/Valentyn1997/CausalTransformer}\\
 \normalsize
 CBRE \cite{CBRE}&\normalsize IHDP,Twins,Jobs&\normalsize Tensorflow&\url{https://github.com/jameszhou-gl/CBRE}\\
 \normalsize
 CITE \cite{CITE}&\normalsize IHDP,Twins,Jobs&\normalsize Tensorflow&\url{https://github.com/XinshuLI2022/CITE}\\
 \normalsize
 ADMIT \cite{ADMIT}&\normalsize TCGA,News&\normalsize Pytorch&\url{https://github.com/waxin/ADMIT}\\
  \bottomrule
\end{tabular}
\end{threeparttable}
}
\end{center}
\end{table*}

By combining related methods, datasets, and source codes, we might more easily identify the innovation points in each model. Meanwhile, it will also facilitate more fair comparison in performance evaluation. In addition, undoubtedly, these source codes will also greatly promote the development of research community about causal effect estimation. As an example, the covariate decomposition is applied to the Dragonnet \cite{Dragonnet} model when combined with the DeR-CFR \cite{DeR-CFR} model to make a further model optimization. By applying the TransTEE \cite{TransTEE} attention mechanism to the representation balance part of VCNet \cite{VCNet} or DRNet \cite{DRNet}, the continuous treatment estimation curve might be fitted more accurately. It also means the latest advances in causal analysis have also benefited from or inspired by some previous representative works.

\section{Conclusions and Future Prospect in Industrial Application}
\label{sec:Conclusions}
\subsection{Conclusions}
Deep causal models have become increasingly popular as a research topic because of the development of causal effect estimation and deep learning. It is possible to improve causal effect estimation accuracy and unbiasedness by applying deep network models to causal effect estimation. Moreover, the deep network might be optimized and improved by the profound theory used in causal effect estimation. This survey presents the development of deep causal models and the evolution of various methods. Firstly, the basic knowledge related to the field of causal effect estimation is adhibited. Then, we present the classical treatments and metrics. Additionally, we provide a comprehensive analysis of the deep causal model from temporal development. Next, we divide the deep causal modeling methods into five groups with an overview and analysis. Furthermore, We furnish a comprehensive conclusion of the application of causal effect estimation in industry. Finally, we summarize the relevant benchmark datasets, open source codes, and performance results as experimental guidelines. 

Since 2016, causal effect estimation has been combined with deep learning models for the first time in the binary treatment scenario for estimation of counterfactual outcomes. So far, deep causal models have been used for time-series, multivariate treatment, and continuous dose treatment situations. It is inseparable from the proposal of deep network models such as AE, GAN, RNN, and Transformer by researchers in the field of deep learning, the generation and simulation of datasets such as IHDP, Twins, Jobs, News, and TCGA by researchers in the field of statistics, and the exploration of ATE, PEHE, MISE, DPE by researchers in industry guided with the theory of Potential Outcome Framework. We believe that with the joint efforts of everyone in the community of causal learning, deep causal models will flourish for the benefit of society and humanity. 
\subsection{Future prospect in industrial application}

For marketing applications, in addition to incentive efficacy evaluation scores, the interpretability of deep learning is also necessary to understand why we predict the output and use it as a basis for business innovation. Furthermore, marketing decisions may involve ethical and legal issues, such as loan applications. Therefore, it makes sense to use causal effect estimation as a guarantee of fairness.

For e-commerce applications, existing debias methods in recommendation systems are familiarly designed to address only one or two specific biases. There is an urgent need for a universal debias framework to deal with all kinds of prejudice. In addition, how to evaluate a recommendation system fairly and impartially is also an important issue. Existing methods either require accurate propensity scores or rely on unbiased data. Therefore, deep causal models are urgently needed to provide theoretical assurance. 

For financial and economic applications, it is desirable to integrate both micro- and macro-level data to study causal effects for stability. In addition, uncertainty quantification is also beneficial to decision making processes, which might be realized by Bayesian approximation and ensemble learning techniques with deep learning.

For medical and educational applications, deep causal models might be adopted in a variety of fields including processing high-dimensional data, enriching randomized trials with real-world data, evaluating spillover causal effects, and moving from studies conducted on specific populations to other populations of interest.


\bibliography{sn-bibliography}% common bib file
%% if required, the content of .bbl file might be included here once bbl is generated


{\section{Appendix}}
\label{sec:Appendix}

\subsection{Treatments and Metrics}
\label{sec:Examples}

This section provides an analysis and description of the disparity performance metrics adopted for disparity classical application scenarios. In addition to the basic metrics in survey \cite{Survey}, we expand the evaluation from binary to multiple and continuous scenarios. 

\subsubsection{Binary treatment}
Treatment refers to the action of a sample or a subject. In the medical field, a medication regimen for a patient is a treatment. When there are only two treatment options, the group of units applied with treatment $T = 1$ is the treated group, and the group of units with $T = 0$ is the control group, which might be referred to as the binary treatment \cite{Survey}.

In the binary treatment situation, the most basic and common performance metric is \textbf{Average Treatment Effect(ATE)} \cite{ATE}, which is defined as:
\begin{equation}
    \text{ATE}  = \mathbb{E}[Y(T = 1) - Y(T = 0)],
\end{equation}
where $Y(T = 1)$ and $Y(T = 0)$ indicate the results of the treatment and control groups in the population. This metric is commonly used to estimate causal effects for the well-known dataset IHDP \cite{IHDP}. Differentiate between treatment and control groups by whether intensive high-quality child care is applied, and the outcome is a score on the cognitive test for the infant. 

In a sample set, the treatment effect is called the \textbf{Conditional Average Treatment Effect (CATE)} \cite{Survey} that is given as follows: 
\begin{equation}
    \text{CATE}  = \mathbb{E}[Y(T = 1) \mid X = n] - \mathbb{E}[Y(T = 0) \mid X = n],
\end{equation}
where $Y(T=1) \mid X=n$ and $Y(T=0) \mid X=n$ represent the results under the sample set for the treatment and control groups with $X=n$, respectively. Since disparity treatments have disparity effects on variation example sets, CATE is also known as heterogeneous treatment effect. 
The treatment effect at the individual level is called \textbf{Individual Treatment Effect (ITE)} \cite{Survey}, which is defined as:
\begin{equation}
    \text{ITE}_n = Y_n(T = 1) - Y_n(T = 0),
    \label{eqn: ITE}
\end{equation}
where $Y_n(T = 1)$ and $Y_n(T = 0)$ represent the results of the sample in the treatment and control groups. 

It is also helpful to keep in mind another evaluation metric, which is called \textbf{Precision in Estimation of Heterogeneous(PEHE)} \cite{IHDP_2}. Demonstrating the capacity to capture heterogeneity in treatment effects at the individual level :
\begin{equation}
\text{PEHE}=\frac{1}{N} \sum_{n=1}^{N}\left(Y^F_{1}\left(n\right)-Y^F_{0}\left(n\right)-\left(Y^{CF}_{1}\left(n\right)-Y^{CF}_{0}\left(n\right)\right)\right)^{2}
\end{equation}
where $Y^F_{1}(n)$, $Y^F_{0}(n)$ and $Y^{CF}_{1}(n)$, $Y^{CF}_{0}(n)$ respectively indicate unbiased estimates of the factual and counterfactual outcome of the n-th sample in the treatment and control groups. The Twins dataset \cite{CEVAE} is typical of the application of this metric, and the result is a one-year mortality rate for children.

For the treated group, the treatment effect is referred to as the \textbf{Average Treatment effect on the Treated group (ATT)} \cite{Survey}, which is defined as:
\begin{equation}
    \text{ATT}  = \mathbb{E}[Y(T = 1) \mid T = 1] - \mathbb{E}[Y(T = 0) \mid T = 1 ],
\end{equation}
where $Y(T = 1) \mid T = 1$ and  $Y(T = 0) \mid T = 1$ correspond to potential treated and control outcome of the treated group respectively. This metric is available for the jobs dataset \cite{Jobs}, treatment group take part in vocational training, while the control group are not and the outcome is employment status.

Since only factual data is available for Jobs dataset, the testing set is from RCT.Therefore, heterogeneity effect indicators only might be described by \textbf{Policy Risk} $\left(\mathcal{R}_{p o l}(\pi)\right)$ \cite{CFRNet}, which is defined as:
\begin{equation}
% \resizebox{.9\hsize}{!}{ R_{\text {pol }}(\pi)=\frac{1}{N} \sum_{n=1}^{N}\left[1-\left(\sum_{i=1}^{K}\left[\frac{1}{\left|\Pi_{i} \cap T_{i} \cap E\right|} \sum_{X(n)\in \Pi_{i} \cap T_{i} \cap E} Y^F_{i}(n) \times \frac{\left|\Pi_{i} \cap E\right|}{|E|}\right]\right)\right] }
\resizebox{.9\hsize}{!}{ $R_{\text {pol }}(\pi)=\frac{1}{N} \sum_{n=1}^{N}\left[1-\left(\sum_{i=1}^{K}\left[\frac{1}{\mid\Pi_{i} \cap T_{i} \cap E\mid} \sum_{X(n)\in \Pi_{i} \cap T_{i} \cap E} Y^F_{i}(n) \times \frac{\mid\Pi_{i} \cap E\mid}{\mid E \mid} \right]\right)\right]$ }
\end{equation}
where $\Pi_{i}=\{X(n): i=\arg \max \hat{Y^{CF}}\}, T_{i}=\left\{X(n): t_{i}(n)=1\right\}$, and $E$ is the subset of RCT.
\subsubsection{Multiple treatment}
Unlike the binary treatment problem, the multiple treatment problem might be described by the following stereotype \cite{PM}: For each sample, the potential outcomes are represented as a vector $Y$ with $N$ entries ${{Y_j}^F}$ where each entry corresponds to the outcome
when applying one treatment $T_j$ out of the set of $N$ available treatments with $ j\in \left\{0,1,2,\dots, N-1 \right\}$. The set of available treatments might contain more than two treatments. As training data, samples $k$ and their observed factual outcomes $Y^{F}$ are received when applying one treatment $T_i$ , the other outcomes might not be observed. The purpose is to train a predictive model $\hat{f}$ that is able to estimate the entire potential outcomes vector $\hat{Y}$ with $k$ entries $Y_j$.

With the exception of ATE and CATE, an accurate estimate of treatment effect might be determined by using \textbf{Root Mean Square Error (RMSE)} \cite{TECE-VAE} for all subgroups, which is defined as:
\begin{equation}
RMSE=\sqrt{\frac{1}{N} \sum_{n=1}^{N} \frac{1}{\mid T \mid} \sum_{j \in T}\left(Y^T(n, j)-Y^P(n, j)\right)^{2}}
\end{equation}
where $T$ is the set of all possible combinations of treatments, $N$ is the number of observations. $Y^T(n, j)$ displays the true outcome when the $j^{th}$ subset is applied to compute the $n^{th}$ observation, whereas $Y^P(n, j)$ displays the predicted outcome when the $j^{th}$ subset is applied to compute the $n^{th}$ observation. The \textbf{Absolute Error} might be calculated in the same way. This metric is commonly used for the News \cite{PM} dataset, which the treatments are the choice of viewing tools, such as smartphones, tablets, desktops, and TVs.

To measure the average of the RMSE between the actual and estimated difference between ITE for each treatment with no treatment ITE, We might incorporate multiple treatment into one calculation criterion, which is called \textbf{Average PEHE} \cite{TECE-VAE}:
\begin{equation} 
\resizebox{.9\hsize}{!}{ $\mathrm{Average PEHE}_{j}={\frac{1}{\mid T \mid}}\sqrt{\frac{1}{N} \sum_{n=1}^{N}\left(\left(Y^T(n, j)-Y^T(n, 0)\right)-\left(Y^{P}(n, j)-Y^{P}(n, 0)\right)\right)^{2}}, j \in(T-T_0)$ }
\end{equation}
where $T_0$ represents the sample set with no treatment applied. It is worth mentioning that the latest research on estimating the effect of multi-causal treatment COVID-19 \cite{COVID_2}. Average PEHE typically serve as performance metrics in dataset CHESS \cite{COVID_1}.
\subsubsection{Continuous treatment}
Vary from binary and multivariate treatment issues, the continuous treatment \cite{SCIGAN} issue might be formulated as follows: Consider to receive observations of the form $(\mathbf{x}^i, T_f^i, Y_f^i)$ for $i = 1, ..., N$, where, for each $i$, these are independent realizations of the random variables $(\mathbf{X}, T_f, Y_f)$, refer to $\mathbf{X}$ as the feature vector lying in some feature space $\mathcal{X}$, containing pre-treatment covariates. The treatment random variable, $T_f$, is in fact a pair of values $T_f = (W_f, D_f)$ where $W_f \in \mathcal{W}$ corresponds to the {\em type} of treatment being administered which lies in the discrete space of $N$ treatments, $\mathcal{W} = \{w_1, ..., w_k\}$, and $D_f$ corresponds to the {\em dosage} of the treatment, which, for a given treatment $w$ lies in the corresponding treatment's dosage space, $\mathcal{D}_w$ (e.g. the interval $[0, 1]$). The set of all treatment-dosage pairs might be defined as $\mathcal{T} = \{(w, d) : w \in \mathcal{W}, d \in \mathcal{D}_w\}$.

In continuous treatment situation, the sample dose-response curve is adopted as the metric. Furthermore, the metric on the test set is different \cite{DRNet}. Therefore, we might use \textbf{Mean Integral Squared Error (MISE)} \cite{DRNet} to measure the accuracy of the estimation of the patient treatment effect on the dose space, which is defined as:
\begin{equation}
\mathrm{MISE}=\sqrt{\frac{1}{K} \frac{1}{N} \sum_{T_n \in \mathcal{T}} \sum_{n=1}^{N} \int_{\mathcal{D}_{T_n}}\left(Y_n(T_n, u)-\hat{Y_n}(T_n, u)\right)^{2} \mathrm{~d} u}
\end{equation}
where $T$ is the set of treatments in the space, $K$ is the number of samples, and $u$ is the treatment dose taken. For a given treatment, $T_n$ lies within the dose space ${D}_{T_n}$. Moreover, $Y_n(T_n, u)$ and $\hat{Y_n}(T_n, u)$ denote the results predicted by the model and the results at the optimal treatment dose, respectively.

In addition, the \textbf{Mean Dose Policy Error (DPE)} \cite{DRNet} is another significant metric of a model's ability to predict the optimal dose point for each individual treatment, which might be defined by:
\begin{equation}
\mathrm{DPE}=\sqrt{\frac{1}{K} \frac{1}{N} \sum_{T_n \in \mathcal{T}} \sum_{n=1}^{N}\left(Y_n\left(T_n, D^{*}_{T_n}\right)-Y_n\left(T_n, \hat{D}^{*}_{T_n}\right)\right)^{2}}
\end{equation}
where, $D^{*}_{T_n}$ and $\hat{D}^{*}_{T_n}$ represent the true optimal dose and the model-determined optimal dose under a treatment, respectively. With \textbf{Sequential Least Squares Estimation}, the optimal dose point for the model might be determined.

In order to compare the optimal treatment dose pair selected by the model with the true optimal treatment dose pair, the mean \textbf{policy error (PE)} \cite{DRNet} is given as:
\begin{equation}
\mathrm{PE}=\sqrt{\frac{1}{N} \sum_{n=1}^{N}\left(Y_n\left(T_n^{*}, D^{*}_{T_n}\right)-Y_n\left(\hat{T_n^{*}}, \hat{D}^{*}_{T_n}\right)\right)^{2}}
\end{equation}
where $T_n^{*}$ and $\hat{T_n^{*}}$ represent the optimal treatment and the optimal treatment determined by the model, respectively. By calculating the optimal dose for each treatment and then selecting the treatment that yields that optimal dose, the optimal dose pair for the model might be selected. The above metrics are typically used in the TCGA \cite{TCGA} dataset. Among them, drug therapy, chemotherapy, and surgery are the treatment options, and the outcome is a risk of cancer recurrence after treatment.

\subsection{Experiment}
We also provide an experimental summary of the binary treatment, multiple treatment and continuous treatment scenarios on the above datasets, respectively. it should be pointed out that the reported results are under unified dataset setting for the compared models. Detailed experimental results and performance comparisons might be found in Tables \ref{tab:IHDP}, \ref{tab:Twins}, \ref{tab:Jobs}, and \ref{tab:multiple}, respectively, in which the \textbf{Mean $\pm$ Std} of the estimated treatment effect results (lower better) are presented. {Since the factual and counterfactual outcome distributions are known for the IHDP \cite{IHDP} dataset, the absolute error in average intervention effects is denoted as $\epsilon_{ATE}$, and the absolute error in precision of estimation of heterogeneous effects is denoted as $\varepsilon_{PEHE}$. For the Twins \cite{Twins} dataset, although both factual and counterfactual outcomes can be observed simultaneously, the underlying distribution is unknown; hence, the absolute errors in average intervention effects and precision of estimation of heterogeneous effects are represented by $\hat{\epsilon}_{ATE}$ and $\hat{\varepsilon}_{PEHE}$, respectively. In the case of the Jobs \cite{Jobs} dataset, since the training set contains only factual data and the test set originates from a randomized controlled trial (RCT), the absolute error in average treatment effects, $\epsilon_{ATT}$, and policy risk $\mathcal{R}_{pol}(\pi_f)$ are used as evaluation metrics \cite{GANITE}.}

\subsubsection{Settings and results for binary treatment}
The study utilized three variation binary treatment datasets, namely IHDP \cite{IHDP}, Jobs \cite{Jobs}, and Twins \cite{Twins}.To obtain the results, the IHDP and Twins datasets were subjected to 10 realizations, with a 63/27/10 split ratio for train/validation/test sets. On the other hand, the Jobs dataset underwent 10 train/validation/test splits with split ratios of 56/24/20. The study presented the experimental results in Tables \ref{tab:IHDP}, \ref{tab:Twins} and \ref{tab:Jobs}, showing the average performance across the variation splits and realizations.  Among them, in-sample represents the data results of the training set and validation set, and out-sample represents the data results of the test set.

\textbf{IHDP}: To create an unbalanced treatment group, a subset of the population was intentionally removed before the experiment, resulting in an uneven distribution. The simulated results were obtained using the “A” setting of the NPCI package, and the real effect was calculated using noise-free results, similar to the methodology used in \cite{IHDP_2}. To further explore the impact of increasing imbalance between the original treatment groups, a biased subsample of the IHDP dataset was constructed. This allowed for a more nuanced analysis of the effects of growing imbalance, adding to the understanding of the overall impact on the results of the experiment.

\textbf{Twins}: The study's dataset focused on same-sex twin couples weighing less than 2,000 grams, resulting in 5,409 records after removing those with missing features. To create a selection bias, the study selected one twin to observe and hid the other. This was done using the procedure $T_i \mid \mathbf{x}_i \sim \operatorname{Bern}\left(\operatorname{Sigmoid}\left(\mathbf{w}^T \mathbf{x}+n\right)\right)$, where  $\mathbf{w}^T$ was randomly chosen from a uniform distribution $(-0.1, 0.1)$ with $40 \times 1$ size, and $\mathcal{N}$ was added noise from a normal distribution with mean 0 and standard deviation 0.1.

\textbf{Jobs}: The Jobs dataset is designed for binary classification, with the aim of predicting unemployment based on a set of features described in \cite{propensity_jobs}. The dataset includes a LaLonde trial sample \cite{matching_jobs} with 297 treatment and 425 control observations, as well as a PSID comparison group with 2490 control observations. At the end of the study, the dataset contained 482 unemployed subjects, accounting for 15$\%$ of the total population.

\begin{table*}[ht]
\tiny
\centering
\renewcommand\arraystretch{1.3}
\renewcommand\tabcolsep{5pt} 
\caption{\textbf{Overview of performance comparisons on IHDP dataset by some representative models in the scenario of binary treatment.}}
\label{tab:IHDP}
% \begin{center}
% \vspace{0.2cm}【
% \resizebox{\textwidth}{!}{
% \begin{threeparttable} 
\begin{tabular}{cccccc}
\toprule
\multirow{2}{*}{\textbf{Methods}} & \multicolumn{2}{c}{\textbf{IHDP$\left(\epsilon_{ATE}\right)$}} &
\multicolumn{2}{c}{\textbf{IHDP($\sqrt{{\varepsilon_{PEHE}}}$)}} &  
\\
& \multicolumn{1}{c}{In-sample} & \multicolumn{1}{c}{Out-sample} &
\multicolumn{1}{c}{In-sample} & \multicolumn{1}{c}{Out-sample} &
\\
\midrule
BNN \cite{BNN,CFRNet} & $0.37 \pm 0.03$ & $0.42 \pm 0.03$ & $2.2 \pm 0.1$ & $2.1 \pm 0.1$\\
TARNet \cite{CFRNet} & $0.26 \pm 0.01$ & $0.28 \pm 0.01$ & $0.88 \pm 0.02$ & $0.95 \pm 0.02$\\
CFR$_{MMD}$ \cite{CFRNet} & $0.30 \pm 0.01$ & $0.31 \pm 0.01$ & $0.73 \pm 0.01$ & $0.78 \pm 0.02$\\
CFR$_{WASS}$ \cite{CFRNet} & $0.25 \pm 0.01$ & $0.27 \pm 0.01$ & $0.71 \pm 0.02$ & $0.76 \pm 0.02$\\
CEVAE \cite{CEVAE} & $0.34 \pm 0.01$ & $0.46 \pm 0.02$ & $2.7 \pm 0.1$ & $2.6 \pm 0.1$\\
GANITE \cite{GANITE} & $0.43 \pm 0.05$ & $0.49 \pm 0.05$ & $1.9 \pm 0.4$ & $2.4 \pm 0.4$\\
SITE \cite{SITE} & \underline{~~~~~~~~~} & \underline{~~~~~~~~~} & $0.604 \pm 0.093$ & $0.656 \pm 0.108$\\
ACE \cite{ACE} & \underline{~~~~~~~~~} & \underline{~~~~~~~~~} & $0.489 \pm 0.046$ & $0.541 \pm 0.061$\\
DKLITE \cite{DKLITE} & \underline{~~~~~~~~~} & \underline{~~~~~~~~~} & $0.52 \pm 0.02$ & $0.65 \pm 0.03$\\
DR-CFR \cite{DR-CFR} & $0.240 \pm 0.032$ & $0.261 \pm 0.036$ & $0.657 \pm 0.028$ & $0.789 \pm 0.091$\\
CETransformer \cite{CETransformer} & \underline{~~~~~~~~~} & \underline{~~~~~~~~~} & $0.46 \pm 0.02$ & \textcolor{red}{$0.51 \pm 0.03$}\\
DeR-CFR \cite{DeR-CFR} & $0.130 \pm 0.020$ & $0.147 \pm 0.022$ & \textcolor{red}{$0.444 \pm 0.020$} & $0.529 \pm 0.068$\\
DONUT \cite{DONUT} & $0.13 \pm 0.01$ & $0.19 \pm 0.02$ & \underline{~~~~~~~~~} & \underline{~~~~~~~~~}\\
NESTER \cite{NESTER} & \textcolor{red}{$0.06 \pm 0.04$} & $0.09 \pm 0.07$ & $0.73 \pm 0.19$ & $0.76 \pm 0.20$ &\\
CBRE \cite{CBRE} & $0.10 \pm 0.01$ & $0.13 \pm 0.02$ & $0.52 \pm 0.00$ & $0.60 \pm  0.1$ \\
CITE \cite{CITE} & $0.09 \pm 0.01$ & \textcolor{red}{$0.11 \pm 0.02$} & $0.58 \pm 0.1$ & $0.60 \pm 0.1$ \\
\bottomrule
\end{tabular}
% \end{threeparttable}
% }
% \end{center}
\end{table*}

\begin{table*}[ht]
\tiny
\centering
\renewcommand\arraystretch{1.3}
\renewcommand\tabcolsep{5pt} 
\caption{\textbf{Overview of performance comparisons on Twins dataset by some representative models in the scenario of binary treatment}}
\label{tab:Twins}
%\begin{center}
%\vspace{0.2cm}
%\resizebox{\textwidth}{!}{
%\begin{threeparttable} 
\begin{tabular}{cccccc}
\toprule
\multirow{2}{*}{\textbf{Methods}} &  \multicolumn{2}{c}{\textbf{Twins$\left(\hat{\epsilon}_{ATE}\right)$}} &
\multicolumn{2}{c}{\textbf{Twins(${\sqrt{\hat{\varepsilon}_{PEHE}}}$)}} &
\\
& \multicolumn{1}{c}{In-sample} & \multicolumn{1}{c}{Out-sample} & \multicolumn{1}{c}{In-sample} & \multicolumn{1}{c}{Out-sample} \\
\midrule
BNN \cite{BNN,CFRNet} & $0.0056 \pm 0.0032$ & $0.0203 \pm 0.0071$ & $0.307 \pm 0.001$ & $0.309 \pm 0.004$ \\
TARNet \cite{CFRNet} & $0.0108 \pm 0.0017$ & $0.0151 \pm 0.0018$ & $0.314 \pm 0.001$ & $0.313 \pm 0.002$ \\
CFR$_{MMD}$ \cite{CFRNet} & \underline{~~~~~~~~~} & \underline{~~~~~~~~~} & $0.312 \pm 0.001$ & $0.316 \pm 0.003$ \\
CFR$_{WASS}$ \cite{CFRNet} &$0.0112 \pm 0.0016$ & $0.0284 \pm 0.0032$ & $0.308 \pm 0.001$ & $0.309 \pm 0.003$ \\
GANITE \cite{GANITE} & $0.0058 \pm 0.0017$ & $0.0089 \pm 0.0075$ & $0.289 \pm 0.005$ & $0.297 \pm 0.016$ \\
SITE \cite{SITE} & \underline{~~~~~~~~~} & \underline{~~~~~~~~~} & $0.309 \pm 0.002$ & $0.311 \pm 0.004$ \\
ACE \cite{ACE} & \underline{~~~~~~~~~} & \underline{~~~~~~~~~} & $0.306 \pm 0.000$ & $0.301 \pm 0.002$ \\
DKLITE \cite{DKLITE} & \underline{~~~~~~~~~} & \underline{~~~~~~~~~} & $0.288 \pm 0.001$ & $0.293 \pm 0.003$ \\
CETransformer \cite{CETransformer}& \underline{~~~~~~~~~} & \underline{~~~~~~~~~} & \textcolor{red}{$0.287 \pm 0.001$} & \textcolor{red}{$0.289 \pm 0.002$} \\
DONUT \cite{DONUT} & \textcolor{red}{$0.0025 \pm 0.0016$} & \textcolor{red}{$0.0033 \pm 0.0026$} & \underline{~~~~~~~~~} & \underline{~~~~~~~~~} \\
NESTER \cite{NESTER} & $0.0034 \pm 0.0026$ &$0.063 \pm 0.0033$ & $0.318 \pm 0.002$ & $0.319 \pm 0.000$ &\\
\bottomrule
\end{tabular}
%\end{threeparttable}
%}
%\end{center}
\end{table*}

\begin{table*}[!ht]
\tiny
\centering
\renewcommand\arraystretch{1.3}
\renewcommand\tabcolsep{5pt}
\caption{\textbf{Overview of performance comparisons on Jobs dataset by some representative models in the scenario of binary treatment}}
\label{tab:Jobs}
%\begin{center}
%\vspace{0.2cm}
%\resizebox{\textwidth}{!}{
%\begin{threeparttable} 
\begin{tabular}{cccccc}
\toprule
\multirow{2}{*}{\textbf{Methods}} & 
\multicolumn{2}{c}{\textbf{Jobs$\left(\epsilon_{ATT}\right)$}} &
\multicolumn{2}{c}{\textbf{Jobs($\mathcal{R}_{pol}(\pi_f)$)}} 
\\
& \multicolumn{1}{c}{In-sample} & \multicolumn{1}{c}{Out-sample} & \multicolumn{1}{c}{In-sample} & \multicolumn{1}{c}{Out-sample} \\
\midrule
BNN \cite{BNN,CFRNet} & $0.04 \pm 0.01$ & $0.09 \pm 0.04$ & $0.20 \pm 0.01$ & $0.24 \pm 0.02$\\
TARNet \cite{CFRNet} & $0.05 \pm 0.02$ & $0.11 \pm 0.04$ & $0.17 \pm 0.01$ & $0.21 \pm 0.01$\\
CFR$_{MMD}$ \cite{CFRNet} & $0.04 \pm 0.01$ & $0.08 \pm 0.03$ & $0.18 \pm 0.00$ & $0.21 \pm 0.01$\\
CFR$_{WASS}$ \cite{CFRNet} & $0.04 \pm 0.01$ & $0.09 \pm 0.03$ & $0.17 \pm 0.01$ & $0.21 \pm 0.01$\\
CEVAE \cite{CEVAE} & $0.02 \pm 0.01$ & $0.03 \pm 0.01$ & $0.15 \pm 0.00$ & $0.26 \pm 0.00$  \\
GANITE \cite{GANITE} & $0.01 \pm 0.01$ & $0.06 \pm 0.03$ & $0.13 \pm 0.01$ & $0.14 \pm 0.01$\\
SITE \cite{SITE} & \underline{~~~~~~~~~} & \underline{~~~~~~~~~}  & $0.224 \pm 0.004$ & $0.219 \pm 0.009$\\
ACE \cite{ACE} & \underline{~~~~~~~~~} & \underline{~~~~~~~~~}  & $0.216 \pm 0.005$ & $0.215 \pm 0.009$ \\
DKLITE \cite{DKLITE} & \underline{~~~~~~~~~} & \underline{~~~~~~~~~}  & $0.13 \pm 0.01$ & $0.14 \pm 0.01$ \\
CETransformer \cite{CETransformer} & \underline{~~~~~~~~~} & \underline{~~~~~~~~~}  & \textcolor{red}{$0.12 \pm 0.01$} & \textcolor{red}{$0.13 \pm 0.00$} \\
DONUT \cite{DONUT} & \textcolor{red}{$0.01 \pm 0.00$} & $0.06 \pm 0.05$ & \underline{~~~~~~~~~} & \underline{~~~~~~~~~}\\
SCI \cite{SCI} & \underline{~~~~~~~~~} & \underline{~~~~~~~~~} & $0.204 \pm 0.008$ & $0.225 \pm 0.014$ \\
DeR-CFR \cite{DeR2022} &$0.053 \pm 0.084$ & $0.093 \pm 0.032$ & $0.187 \pm 0.037$ & $0.208 \pm 0.062$\\
NESTER \cite{NESTER} & $0.06 \pm 0.00$ &\textcolor{red}{$0.02 \pm 0.01$} & \underline{~~~~~~~~~}  & \underline{~~~~~~~~~} \\
CBRE \cite{CBRE} & $0.10 \pm 0.03$ &$0.21 \pm 0.07$ & $0.13 \pm 0.00$ & $0.28 \pm 0.00$ &\\
CITE \cite{CITE} & $0.06 \pm 0.02$ &$0.07 \pm 0.03$ & $0.23 \pm 0.02$ & $0.88 \pm 0.0$ &\\
\bottomrule
\end{tabular}
%\end{threeparttable}
%}
%\end{center}
\end{table*}
 
In Table \ref{tab:IHDP}, we observed that on the IHDP dataset, NESTER \cite{NESTER} utilized inductive bias and heuristics to design a multi-head neural network architecture and regularizer which achieved the best performance in terms of the in-sample ATE. In addition, for out-sample Counterfactual CITE \cite{CITE}, self-supervised information hidden within the data was employed. By appropriately leveraging causal prior knowledge, a balance between representativeness and predictiveness was achieved, resulting in optimal outcomes. The covariate representation decoupling system of DeR-CFR \cite{DeR-CFR} played a crucial role in achieving strong performance on the in-sample Prediction Error of PEHE metric. On the other hand, CETransformer \cite{CETransformer} leveraged its impressive self-supervised and representation learning capabilities to efficiently produce desirable outputs on the out-sample data. In contrast, DONUT standed out ATE metric in Table \ref{tab:Twins} of the Twins dataset due to its use of unobserved confounding orthogonalization constraints. Similarly, CETransformer \cite{CETransformer} remained the best performer in terms of the Prediction PEHE. The same applied to the Jobs data set in Table \ref{tab:Jobs}.

\subsubsection{Settings and results for multiple and continous treatments}
In the field of multiple and continuous dose treatments, researchers have relied on three publicly available datasets: TCGA \cite{TCGA}, News \cite{BNN} and MIMIC III \cite{MIMIC}. These datasets were partitioned into 64/16/20\% for training, validation, and testing. Results obtained from experiments conducted using TARNet, TransTEE, and VCNet on the TCGA dataset, as shown in Table \ref{tab:multiple}, were obtained through the re-implementation of the source code.

The experiment \cite{DRGAN,SCIGAN} involved administering three treatments, each with a unique dose and associated with a set of parameters. These parameters were randomly sampled from a normal distribution and normalized. Gaussian noise was added to the results obtained from the experiment. The dose for each treatment was assigned by drawing from a beta distribution, and the treatment assignment was based on a categorical distribution with a softmax function of a $\kappa$ value. The degree of selection bias could be adjusted by changing the $\kappa$ value.

\textbf{TCGA}: The experiment prioritized the 4000 most variable genes and scaled each gene expression feature within the $[0, 1]$ range. To make the treatment and dose more relevant, the experiment categorized them as either chemotherapy, radiotherapy, or immunotherapy. This approach aimed to provide a more interpretable and clinically relevant experimental design \cite{DRNet}.

\textbf{News}: During the experiments, researchers gathered a dataset consisting of 10,000 news articles. Each article was composed of 2,858 features. The two variables, treatment and dose, were used to examine the impact of variation viewing devices and reading times on the articles. Treatment referred to the type of device used to view the article, such as a phone or tablet. Dose referred to the amount of time spent reading the article \cite{CFRNet,DRNet}. 

\textbf{MIMIC}: In this study, 3000 patients who received antibiotics treatment were analyzed. The researchers collected 9 clinical covariates, including age, temperature, heart rate, systolic and diastolic blood pressure, SpO2, FiO2, glucose, and white blood cell count, which were measured at the beginning of the patients' ICU stay. All features were scaled to fit within the range of 0 to 1. The study investigated the impact of variation antibiotics and their doses, which were considered as treatment variables \cite{SCIGAN}.

\begin{table*}[ht]
\tiny
\centering
\renewcommand\arraystretch{1.3}
\renewcommand\tabcolsep{5pt}
\caption{\textbf{Overview of performance comparisons on TCGA dateset by some representative models in the scenario of multiple and continuous treatments}}
\label{tab:multiple}
%\begin{center}
%\vspace{0.2cm}
%\resizebox{\textwidth}{!}{
%\begin{threeparttable} 
\begin{tabular}{ccccc}
\toprule
\multirow{2}{*}{\textbf{Methods}} &
\multicolumn{3}{c}{\textbf{TCGA}} & \\
&
\multicolumn{1}{c}{${\text{MISE}}$} &
\multicolumn{1}{c}{${\text{DPE}}$} &
\multicolumn{1}{c}{${\text{PE}}$} 
\\
\midrule
TARNet \cite{CFRNet} & $5.76 \pm 0.15$ & $0.53 \pm 0.06$ & $0.65 \pm 0.07$  \\
DRN-W \cite{DRNet} & $3.71 \pm 0.12$ & $0.50 \pm 0.05$ & $0.63 \pm 0.05$   \\
DRNet \cite{DRNet} & $3.64\pm 0.12$ & $0.51 \pm 0.05$ & $0.67 \pm 0.05$  \\
DRGAN,SCIGAN \cite{SCIGAN} & \textcolor{red}{$1.89 \pm 0.05$} & $0.31 \pm 0.05$ & \textcolor{red}{$0.25 \pm 0.05$}  \\
VCNet \cite{VCNet} & $6.36 \pm 0.11$ & $0.22 \pm 0.04$ & $0.32 \pm 0.02$ \\
TransTEE \cite{TransTEE} & $6.40 \pm 0.14$ & $0.17 \pm 0.03$ & $0.78 \pm 0.05$ \\
TransTEE+TR \cite{TransTEE} & $6.26 \pm 0.13$ & \textcolor{red}{$0.08 \pm 0.02$} & $0.96 \pm 0.06$ \\
TransTEE+PTR \cite{TransTEE} & $6.36 \pm 0.14$ & $0.18 \pm 0.03$ & $0.81 \pm 0.05$ \\
\bottomrule
\end{tabular}
%\end{threeparttable}
%}
%\end{center}
\end{table*}

In the Table \ref{tab:multiple} of TCGA dataset, DRGAN \cite{DRGAN} and SCIGAN \cite{SCIGAN} showscenario remarkable effectiveness in terms of two key evaluation metrics, MISE and PE. This could be attributed to their innovative architecture, which combined a generator, discriminator, and prediction network. TransTEE \cite{TransTEE} incorporated Transformer representation, propensity score network for treatment effectiveness estimation, and GAN to overcome selection bias, achieving superior performance in the DPE metric.

Based on the experimental analysis, we found that with the development and evolution of the deep model, it has significant implications on solving the core issues and challenges such as representation learning, de-biasing, and counterfactual inference for causal effect estimation. Whether it is in the decision making of therapeutic interventions, the fitting of dose curves or the capture of time-varying confounding, deep causal models give us a new perspective to explore a broader range of directions.

\end{document}